\documentclass[10pt,twocolumn,letterpaper]{article}

\usepackage{cvpr}              
\definecolor{cvprblue}{rgb}{0.21,0.49,0.74}
\usepackage[pagebackref,breaklinks,colorlinks,citecolor=cvprblue]{hyperref}
\usepackage{multirow}
\usepackage{makecell}
\usepackage{pifont}
\usepackage{placeins}
\usepackage{threeparttable}
\usepackage[hang,flushmargin]{footmisc}
\usepackage{wrapfig}

\def\paperID{3895} 
\def\confName{CVPR}
\def\confYear{2026}

\definecolor{darkred}{rgb}{0.7, 0.0, 0.0}
\definecolor{darkred2}{rgb}{0.5, 0.0, 0.0}
\definecolor{darkred3}{rgb}{0.9, 0.0, 0.0}
\definecolor{darkgreen}{rgb}{0.0, 0.42, 0.24}
\definecolor{darkblue}{rgb}{0.10, 0.17, 0.8}
\definecolor{Gray}{gray}{0.93}

\def\eg{\emph{e.g.}} 
\def\ie{\emph{i.e.}} 
\def\etc{\emph{etc.}\xspace}

\newcommand{\method}{PhotoFramer\xspace}
\newcommand{\supp}{Appendix\xspace}

\title{\method: Multi-modal Image Composition Instruction}

\author{
Zhiyuan You$^{12}$, Ke Wang$^3$, He Zhang$^4$, Xin Cai$^2$, Jinjin Gu$^5$,\\Tianfan Xue$^{267\dag}$, Chao Dong$^{168\dag}$, Zhoutong Zhang$^3$\vspace{2pt}\\
$^1$Shenzhen Institutes of Advanced Technology, Chinese Academy of Sciences\\
$^2$Multimedia Laboratory, The Chinese University of Hong Kong\\
$^3$Adobe NextCam \: $^4$Adobe Research \: $^5$INSAIT, Sofia University ``St. Kliment Ohridski''\\
$^6$Shanghai AI Laboratory \: $^7$CPII under InnoHK \: $^8$Shenzhen University of Advanced Technology\\
{\tt\small zhiyuanyou@foxmail.com, tfxue@ie.cuhk.edu.hk, chao.dong@siat.ac.cn, zhoutongz@adobe.com}\\
{\small Project Page: \url{https://zhiyuanyou.github.io/photoframer} \quad\quad $^\dag$ Corresponding Author}
}

\begin{document}

\twocolumn[{
\renewcommand\twocolumn[1][]{#1}
\maketitle
\thispagestyle{empty}
\vspace{-30pt}
\begin{center}
    \includegraphics[width=0.95\linewidth]{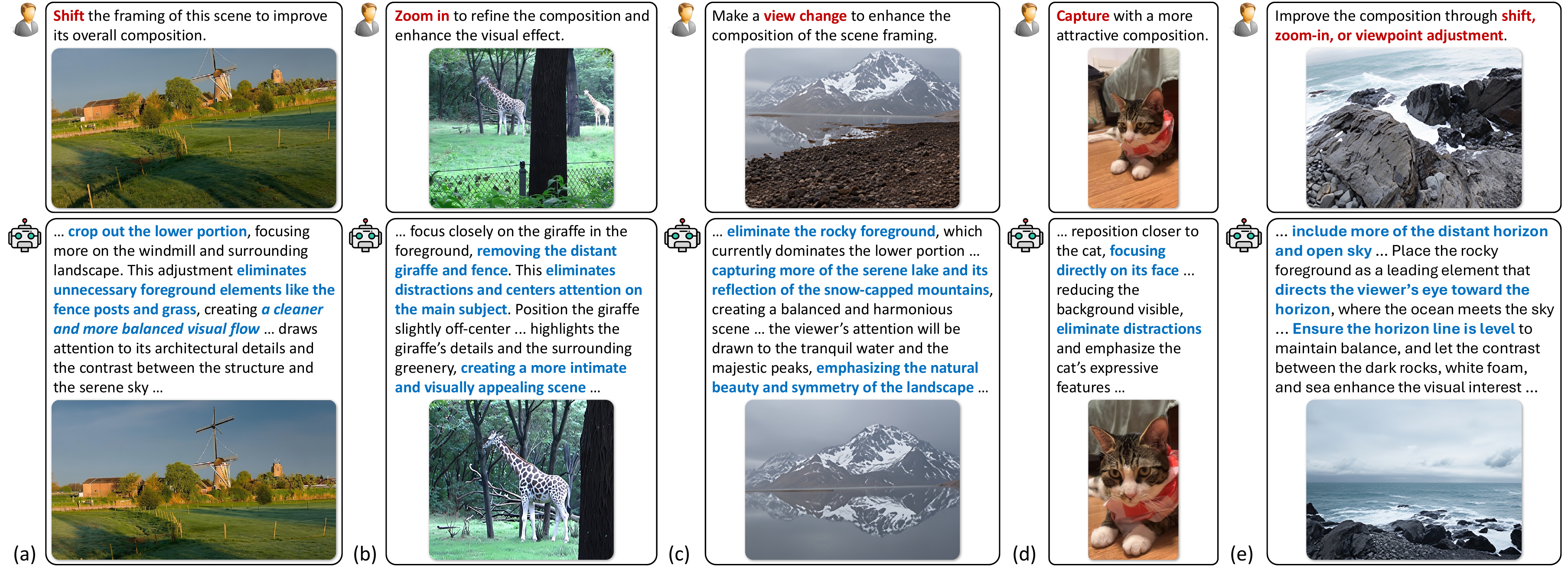}
    \vspace{-12pt}
    \captionof{figure}{
    We propose \method, a model designed for composition instruction during photo capturing. Given a poorly composed image, \method first describes how to improve the composition in natural language and then generates an example image that follows the described suggestions. 
    The photo-taker can follow the textual guidance and the example image to capture a better-composed photo. 
    }
\label{fig:teaser}
\vspace{-2pt}
\end{center}
}]

\vspace{-10pt}
\begin{abstract}

Composition matters during the photo-taking process, yet many casual users struggle to frame well-composed images. 
To provide effective composition guidance, we introduce \method, a multi-modal composition instruction framework. 
Given a poorly composed image, \method first describes how to improve the composition in natural language and then generates a well-composed example image. 
To train such a model, we curate a large-scale dataset. 
Inspired by how humans take photos, we organize composition guidance into a hierarchy of sub-tasks: shift, zoom-in, and view-change tasks. 
Shift and zoom-in data are sampled from existing cropping datasets, while view-change data are obtained via a two-stage pipeline. 
First, we sample pairs with varying viewpoints from multi-view datasets, and train a degradation model to transform well-composed photos into poorly composed ones. 
Second, we apply this degradation model to expert-taken photos to synthesize poor images to form training pairs. 
Using this dataset, we finetune a model that jointly processes and generates both text and images, enabling actionable textual guidance with illustrative examples. 
Extensive experiments demonstrate that textual instructions effectively steer image composition, and coupling them with exemplars yields consistent improvements over exemplar-only baselines. 
\method offers a practical step toward composition assistants that make expert photographic priors accessible to everyday users.

\end{abstract}
\vspace{-10pt}
    
\vspace{-6pt}
\section{Introduction}
\label{sec:intro}
\vspace{-4pt}

Modern mobile cameras have become increasingly powerful~\cite{dxomark, camera1, camera2, camera3}. 
Equipped with these cameras, the users could capture high-resolution, noise-free, and well-exposed photos. 
However, many casual users still struggle to capture visually pleasing photos. 
As shown in the top images of \cref{fig:teaser}, these photos appear visually unappealing due to their poor composition. 
For instance, in \cref{fig:teaser}e, the sea horizon is tilted relative to the image border, and the foreground rocks dominate the frame excessively. 
Therefore, providing amateur photographers with composition guidance during photo capturing is of great importance~\cite{cpam, cadb, retrieval3}.

Casual users can be guided more effectively through a combination of textual and visual instructions. 
Consider a captured farm scene in \cref{fig:teaser}a, textual instruction ``eliminates the fence posts and grass'' provides concrete and actionable operations, along with detailed reasons (\eg, ``creating a cleaner and more balanced visual flow''). 
Meanwhile, in the bottom of \cref{fig:teaser}a, the well-composed example photo capturing the same farm scene is intuitive and easy to follow, consistent with prior findings~\cite{retrieval2, retrieval3}.

To leverage the strengths of both modalities, we introduce \method, a multi-modal image composition instruction framework that provides detailed textual guidance paired with corresponding visual example photos.

To build this composition instruction model, we design a hierarchical set of tasks inspired by how humans take photos. 
Humans typically first determine a suitable position and angle (\ie, vantage point), choose an appropriate focal length, and then finetune subject placement and alignment~\cite{photo_process1, photo_process2}. 
Accordingly, \method consists of three sub-tasks: view-change (\cref{fig:teaser}c), zoom-in (\cref{fig:teaser}b), and shift (\cref{fig:teaser}a) tasks. 
In addition, when the user does not specify a task type (as in \cref{fig:teaser}de), \method automatically determines suitable operations. 
We show that the model does not merely select one task type, but adaptively fuses multiple operations to achieve better composition.

Under this task formulation, we construct a comprehensive multi-modal dataset containing ``poor image, good image, text guidance'' triplets. 
For the shift and zoom-in sub-tasks, we sample image pairs from existing cropping datasets ~\cite{gaic_v2, cpc, flickrcrop, cuhkcrop, sacnet_sacd}. 
For the view-change sub-task, we first sample image pairs with different viewpoints from multi-view datasets~\cite{dl3dv}, and then train a degradation model that converts well-composed photos into compositionally degraded ones. 
We then apply this model to expert-taken photos~\cite{unsplash_lite} to synthesize view-change examples. 
Finally, we employ one vision-language model~\cite{qwen2.5_vl} to annotate text guidance for each pair. 
In total, we collect 207K triplets, serving as the foundation for model training.

Equipped with the above dataset, we finetune the unified understanding-generation model~\cite{nextgpt, seedx, chameleon, showo, bagel} to generate both textual guidance and example image. 
As depicted in \cref{fig:teaser}, our model is required to process inputs and outputs that include both texts and images. 
Consequently, vision-language models~\cite{llava, minigpt4, qwen2.5_vl} that can only produce textual outputs, as well as image generation or editing models~\cite{ldm, kontext, instructmove, cai2025parametric} that can only produce images, are unsuitable. 
We therefore employ the unified understanding-generation model Bagel~\cite{bagel} as our base and fine-tune it using our proposed dataset. 
Experiments show that textual guidance effectively guides image generation, highlighting the advantages of the text-vision joint modeling framework.

Extensive comparisons and ablations further confirm the superiority of our \method over baselines. 
We hope this work serves as a stepping stone toward composition assistants that help everyday users capture expert-level photos.

\vspace{-4pt}
\section{Related Works}
\label{sec:related_works}
\vspace{-2pt}

\textbf{Composition understanding} is the foundation for composition instruction. 
Composition classification~\cite{kupcp, cadb, picd} defines multiple composition categories (\eg, rule-of-thirds, symmetrical, \etc). 
Some works~\cite{cadb, gaic_v1, gaic_v2} focus on score-based composition assessment. 
Recent works~\cite{artimuse, aesr1} utilize powerful vision language models~\cite{internvl3, qwen2.5_vl, gpt4v} to build multi-aspect aesthetics (\ie, including composition) assessment models with joint scoring and descriptive outputs.

\vspace{2pt}\noindent\textbf{Image cropping} is a common approach to enhance the composition. Given an original image, this task aims to find a cropped patch with better composition. 
Extensive cropping datasets can be broadly categorized into two types. 
The first type is densely annotated, where multiple crops are labeled per image~\cite{cpc, gaic_v1, gaic_v2, sacnet_sacd, s2cnet}. 
The second type is sparsely annotated, where only the best crop is labeled per image~\cite{cuhkcrop, flickrcrop, flms}. 
Built upon these datasets, deep learning models have achieved substantial progress. 
Many methods~\cite{cpc, gaic_v1, gaic_v2, sacnet_sacd, s2cnet, cuhkcrop, flms, diverse_global, human_centric, propose_select_1, propose_select_2, propose_select_3} adopt a two-stage strategy: first generating crop candidates, then selecting the optimal one. 
Other coordinate regression methods~\cite{vfn, reg2, reg3, reg4, reg5, reg6} directly predict crop boundaries. 
Recently, GenCrop~\cite{gencrop} leverages Stable Diffusion~\cite{ldm} to outpaint professional photos for dataset expansion. 
ProCrop~\cite{procrop} retrieves compositionally similar reference images to guide the cropping process. 
However, image cropping remains a post-processing operation applied after photo capture. 
In contrast, our work aims to guide users \textit{during} the capturing process to take well-composed photos.

\vspace{2pt}\noindent\textbf{Composition guidance}. 
Traditional methods rely on retrieval to search similar images in the database as user reference~\cite{retrieval1, retrieval2, retrieval3}. 
However, retrieval-based guidance suffers from scene and subject mismatch, making it hard to follow. 
Recent CPAM~\cite{cpam} could automatically provide photographers with camera pose adjustment guidance. 
Our work differs from CPAM in three key aspects. 
First, CPAM predicts only adjusted yaw and pitch angles, while we generate both text instruction and a good example image, offering a more intuitive and informative guide. 
Second, CPAM is limited to yaw and pitch adjustments, whereas our model additionally supports zoom control and large-scale viewpoint changes. 
Third, CPAM employs separate models for understanding and adjustment, while we adopt a unified model that jointly performs both tasks, enabling mutual enhancement.

\vspace{2pt}\noindent\textbf{Unified multi-modal model} conducts image understanding and generation in a unified model~\cite{emu, nextgpt, seedx, chameleon, metamorph, janus, showo, bagel}. 
Since our aim is to output both textual instructions (\ie, understanding) and example images (\ie, generation), we take the unified multi-modal model as our base model.

\vspace{-5pt}
\section{Task Paradigm and Dataset Construction}
\label{sec:data}
\vspace{-2pt}

\begin{figure*}[tb]
    \centering
    \includegraphics[width=0.95\linewidth]{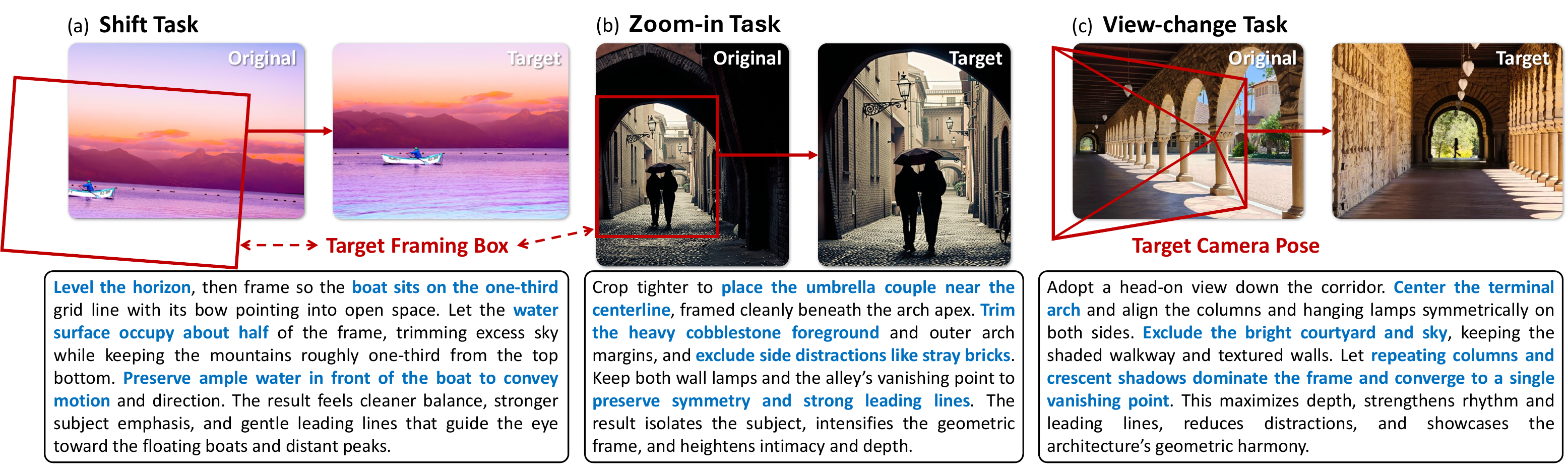}
    \vspace{-10pt}
    \caption{
    Task paradigm and data example. Given a poorly composed image, our \method is required to generate a \textit{textual guidance} (describing how to improve the composition) together with an \textit{example image} (depicting what a well-composed image looks like). 
    Motivated by three key photography factors (\textit{vantage point}, \textit{focal choice}, and \textit{subject placement}), our \method comprises three tasks:
    (a) \textbf{Shift}: adjust the framing to place the subject properly and remove border distractions;
    (b) \textbf{Zoom-in}: select a tighter crop (simulating a longer focal length) that yields a stronger composition;
    (c) \textbf{View-change}: choose a new vantage point or camera pose to reframe the scene.
    }
    \label{fig:task}
    \vspace{-17pt}
\end{figure*}

\subsection{Task Paradigm}
\label{subsec:task}
\vspace{-2pt}

We first construct the task paradigm for composition guidance. 
Our goal is to develop an assistant that can guide humans during the photo-taking process. 
Therefore, we begin by revisiting how humans take photos and analyzing the key abilities the model should possess. 
We summarize three key steps to decompose the capturing process from~\cite{photographereye}.

\noindent-- First, humans choose an appropriate shooting position and angle, referred to as the \textit{vantage point}. As shown in \cref{fig:task}c, given a scene, our model needs to infer alternative viewpoints and select the one with the best composition.

\noindent-- Second, humans adjust the focal length (or zoom level on mobile photo cameras) to emphasize specific subjects. Accordingly, \method should have the ability to identify well-composed crops within a larger scene (\cref{fig:task}b).

\noindent-- Third, humans carefully adjust the camera to maintain a level frame, avoid border distractions, and place subjects in balanced positions. Thus, as shown in \cref{fig:task}a, \method should position subjects appropriately (\eg, centered or rule of thirds) and remove distractions to maintain clean borders.

Based on the above discussion, as illustrated in \cref{fig:task}, we design a hierarchical task paradigm that progressively guides our \method to acquire these capabilities:

\noindent-- \textit{Shift task}. Given a poorly composed image, \method adjusts the framing to properly place the subject, levels the image, and removes border distractions.

\noindent-- \textit{Zoom-in task}. Given an original image, \method generates a tighter crop with improved composition.

\noindent-- \textit{View-change task}. Given a captured scene, \method selects a new vantage point or camera pose to reframe the scene and generates the corresponding image.

Moreover, both textual guidance and example images are important. 
Example images are intuitive and easy to follow, while the textual guidance provides detailed reasoning for the generated image and is easier to understand. 
Therefore, as depicted in \cref{fig:task}, our \method\ is designed to generate a detailed \textit{text guidance} (describing how to improve the composition) together with an \textit{example image} (demonstrating what a well-composed image should look like).

Formally, let the input poor-composition image be denoted as \texttt{I\_poor}, the task type (\ie, expressed in text) as \texttt{T\_task}, the generated text guidance as \texttt{T\_guide}, and the target well-composed image as \texttt{I\_good}. 
Our \method, denoted as \texttt{f()}, is trained to perform the following mapping: \texttt{I\_good, T\_guide = f(I\_poor, T\_task)}.

\vspace{-3pt}
\subsection{Dataset Construction}
\vspace{-3pt}

Data is the key factor in training such a unified model. Following~\cite{bagel, metamorph}, we need to construct \texttt{<T\_task,I\_poor, I\_good,T\_guide>} pairs for supervision.

\vspace{2pt}\noindent\textbf{Task prompt collection}. 
For \texttt{T\_task}, we follow~\cite{depictqa, depictqav2, qalign} by predefining some text templates for each task and randomly sampling one template to form the pair. 
For example, a template for the shift task is ``shift the scene to enhance the composition''. See \supp for all task prompts.

\vspace{2pt}\noindent\textbf{Image pair collection}. Collecting \texttt{<I\_poor,I\_good>} image pairs is the key process and will be detailed later.

\vspace{2pt}\noindent\textbf{Text guidance collection}. 
Given \texttt{<I\_poor,I\_good>} image pairs, we employ a vision-language model, Qwen2.5-VL-32B~\cite{qwen2.5_vl}, to generate text guidance \texttt{T\_guide}. 
Specifically, we input the poor and good images along with the task type, and prompt the model to describe how to transform the poor image into the good one, with detailed justifications.

\setlength{\columnsep}{7pt}
\begin{wraptable}{r}{0.28\textwidth}
    \centering
    \footnotesize
    \setlength\tabcolsep{1pt}
    \vspace{-12pt}
    \caption{Dataset statistics.}
    \vspace{-8pt}
    \begin{tabular}{@{}c|ccc@{}}
    \toprule
    & Shift & Zoom-in & View-change \\
    \midrule
    \# Original & 10,321 & 7,665 & \multirow{2}{*}{27,393} \\
    \# Pairs & 164,904 & 14,182 & \\
    \bottomrule
    \end{tabular}
    \label{tab:data_statistic}
    \vspace{-9pt}
\end{wraptable}
\vspace{2pt}\noindent\textbf{Dataset statistics} are given in \cref{tab:data_statistic}. 
Since the shift and zoom-in pairs are sampled from the original images in the cropping datasets (see follows), both the original images and pairs are included in the statistics. 
In total, our constructed dataset comprises 45K original images and 207K pairs, providing a solid foundation for model training.

\vspace{-2pt}
\subsubsection{Shift and Zoom-in Pairs}
\label{subsubsec:shift_zoomin_data}

\begin{figure}[tb]
    \centering
    \includegraphics[width=0.95\linewidth]{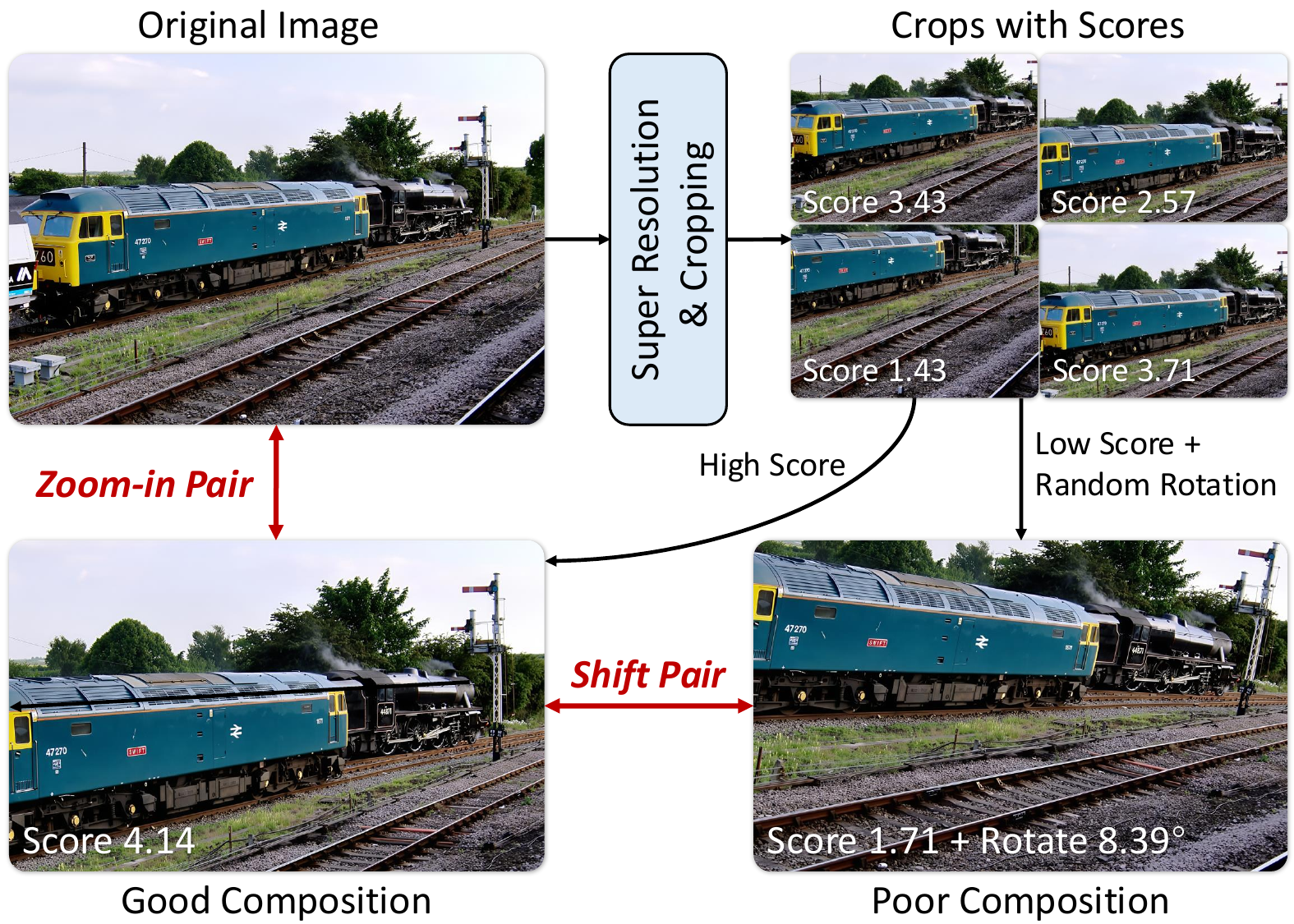}
    \vspace{-8pt}
    \caption{
    Dataset construction for the shift and zoom-in tasks. 
    For the shift task, given an image from the cropping dataset, we sample its crops to form a \texttt{<poor,good>} image pair. A random rotation is applied to the poor crop. 
    For the zoom-in task, the original image and a well-composed crop form an \texttt{<original,good>} pair. 
    To ensure sufficient resolution, we apply $4\times$ super resolution to the original image using HYPIR~\cite{hypir}. 
    }
    \label{fig:data_shift_zoomin}
    \vspace{-17pt}
\end{figure}

As illustrated in \cref{fig:data_shift_zoomin}, we construct shift and zoom-in pairs from existing cropping datasets, which offers two advantages. 
First, the scenes and subjects in cropping datasets are curated by contributors, making them well-suited for composition-related training. 
Second, they contain useful annotations, which could reduce annotation workload.

\vspace{2pt}\noindent\textbf{Shift pair collection}. We collect shift pairs from existing cropping datasets GAIC~\cite{gaic_v2} and CPC~\cite{cpc}. 
GAIC provides scores for each crop, while CPC contains only selected good and best crops, which cannot be directly used. 
To generate missing scores in CPC, we design a mathematical sampling model in \supp. 
As shown in \cref{fig:data_shift_zoomin}, crops with scores above 4.0\footnote{Composition scores in this work are all normalized to a [1,5] scale.} are treated as good images, while those below 2.0 are treated as poor images, forming \texttt{<I\_poor,I\_good>} pairs. 
A small proportion of mid-score crops are sampled as poor images to enhance robustness. 
Finally, a random rotation is applied to the poor crop for augmentation. 
Note that our work focuses on composition instructions without changing aspect ratio. 
Hence, we discretize the aspect ratio range [0.45, 2.2] into 11 values, and only images with the same aspect ratio could be paired.

\vspace{2pt}\noindent\textbf{Zoom-in pair collection}. We pair each good crop with its original image to form a \texttt{<I\_poor,I\_good>}\footnote{We use ``poor'' for simplicity. The original image may not be strictly poor.} pair, as depicted in \cref{fig:data_shift_zoomin}. 
Thus, the key step is to identify the good crop. 
We adopt existing cropping datasets including GAIC~\cite{gaic_v2}, CPC~\cite{cpc}, SACD~\cite{sacnet_sacd}, FlickrCrop~\cite{flickrcrop}, FLMS~\cite{flms}, and CUHKCrop~\cite{cuhkcrop}. 
For GAIC and CPC, where a score can be assigned to each crop, crops with scores above 4.0 are regarded as good crops. 
Other datasets provide a human-labeled best crop, which can be directly used. 
To ensure the two paired images share the same aspect ratio, we crop the original image with the largest possible region that matches the good crop's aspect ratio. 
The crop center is chosen as close as possible to the image center, subject to fully containing the good crop.

\vspace{2pt}\noindent\textbf{Super-resolution for sufficient resolution}. 
Some crops have very low resolution, \ie, below 300 pixels, which is hard to use. 
Therefore, before cropping, we apply a $4 \times$ super-resolution to the original image using HYPIR~\cite{hypir}.

\begin{table}[t]
\setlength\tabcolsep{1.5pt}
\centering
\footnotesize
    \caption{
    Results of our assessment model. 
    Metrics for assessment and classification tasks are SRCC / PLCC and accuracy. 
    CADB~\cite{cadb} and GAIC~\cite{gaic_v2} datasets are for composition assessment, while AVA~\cite{ava} dataset is for aesthetic assessment. 
    }
    \vspace{-8pt}
    \begin{tabular}{c|ccc|c}
    \toprule
    Task & \multicolumn{3}{c|}{Composition / Aesthetic Assessment} & Classif.  \\
    \midrule
    Dataset & CADB & GAIC & AVA & CADB \\
    \midrule
    VFN~\cite{vfn} & 0.052 / 0.049 & 0.152 / 0.162 & 0.139 / 0.142 & - \\
    VEN~\cite{cpc} & 0.084 / 0.082 & 0.410 / 0.428 & 0.232 / 0.241 & - \\
    AutoPhoto~\cite{autophoto} & 0.065 / 0.079 & 0.407 / 0.427 & 0.604 / 0.613 & - \\
    Q-Align~\cite{qalign} & 0.561 / 0.557 & 0.169 / 0.178 & 0.809 / 0.804 & - \\
    Qwen2.5-VL-32B~\cite{qwen2.5_vl} & 0.420 / 0.426 & 0.195 / 0.205 & 0.527 / 0.492 & 0.101 \\
    \midrule
    Our Model (7B) & \textbf{0.763} / \textbf{0.777} & \textbf{0.795} / \textbf{0.805} & \textbf{0.825} / \textbf{0.828} & \textbf{0.583} \\
    \bottomrule
    \end{tabular}
\label{tab:res_assess}
\hfill
\vspace{3pt}
    \centering
    \includegraphics[width=1.0\linewidth]{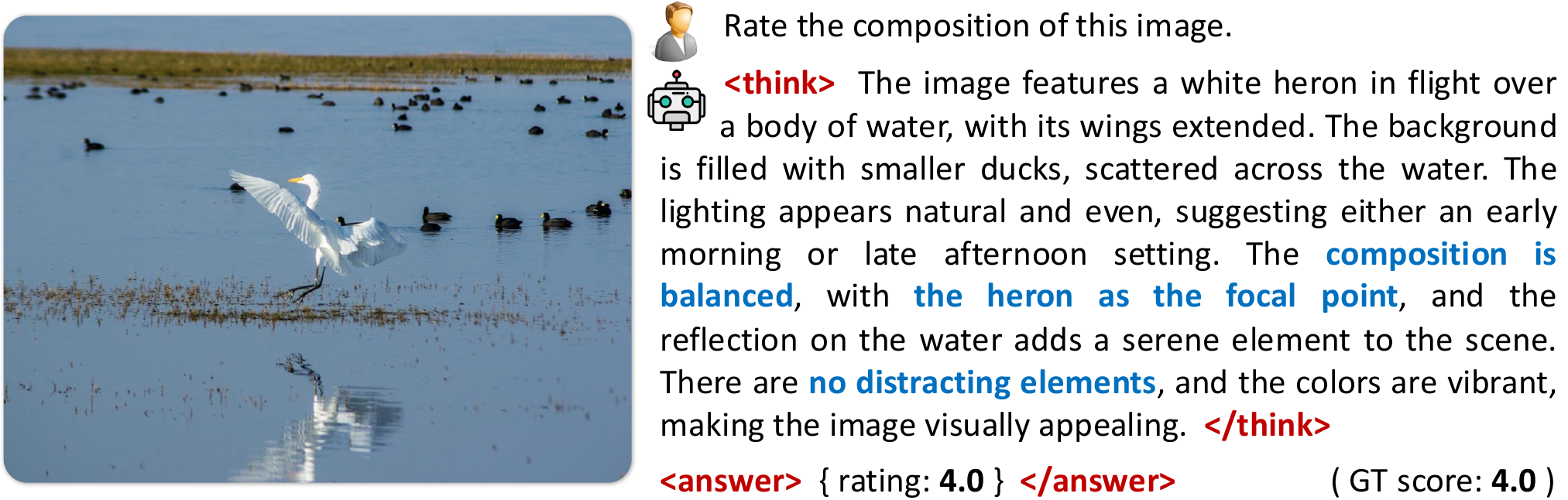}
    \vspace{-18pt}
    \captionof{figure}{
    Qualitative results of our composition assessment model, illustrating the thinking process and final assessment output.
    }
    \label{fig:res_assess}
    \vspace{-15pt}
\end{table}

\begin{figure*}[tb]
    \centering
    \includegraphics[width=0.95\linewidth]{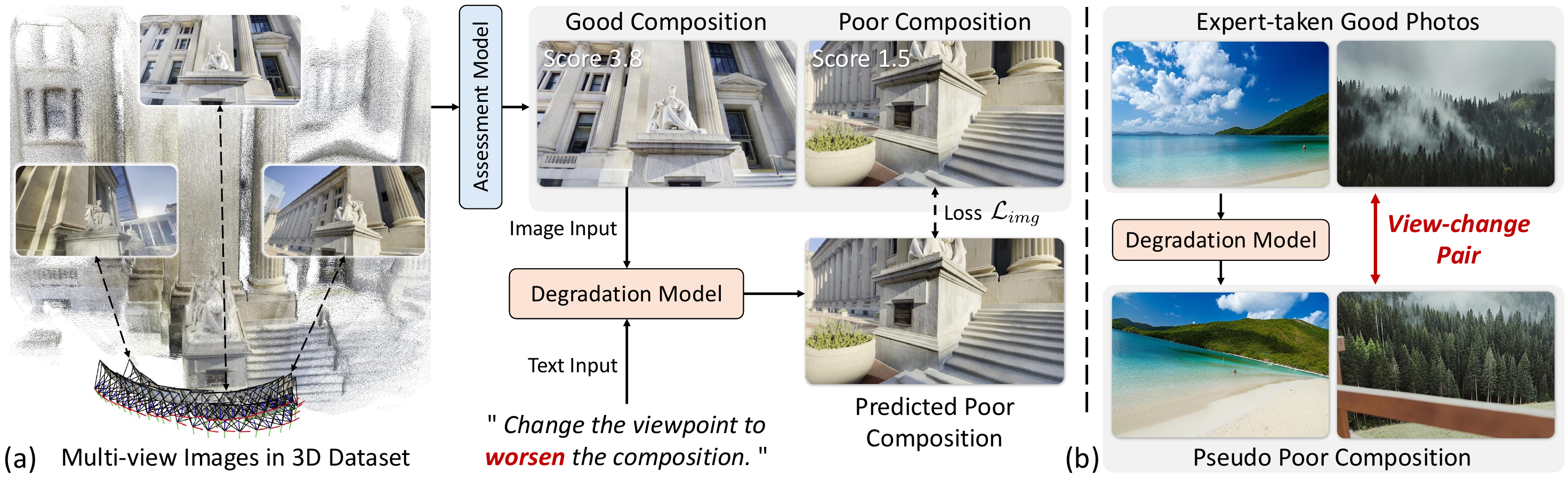}
    \vspace{-10pt}
    \caption{
    Dataset construction for the view-change task. 
    (a) Leveraging our composition assessment model in \cref{subsubsec:assess_model}, we sample \texttt{<poor,good>} image pairs from multi-view datasets. We then train a composition degradation model that generates poor-composition images from good ones. 
    (b) We apply this degradation model to expert-taken good photos to synthesize pseudo poor-composition images, forming the final pairs. 
    We do not rely solely on multi-view datasets, as most of their good images are not sufficiently well-composed. 
    }
    \label{fig:data_view}
    \vspace{-15pt}
\end{figure*}

\vspace{2pt}\noindent\textbf{Image pair filtering}. 
Three types of filtering are performed to ensure data quality. 
(1) Pairs containing any image with an aspect ratio outside [0.45, 2.2] are discarded. 
(2) For shift pairs, we remove pairs depicting different subjects. 
(a) Pairs with CLIP similarity~\cite{clip} below 0.8 are discarded. 
(b) We use U$^2$Net~\cite{u2net} to extract saliency masks of the subjects, compute DINOv2~\cite{dinov2} features on the masked regions, and discard pairs with subject-level cosine similarity below 0.6. 
(c) To avoid confusion with the zoom-in task, we filter out pairs where one image is fully contained in the other and the area ratio is below 0.6. 
(d) To ensure sufficient composition difference in each pair, we retain those pairs whose good score exceeds the poor score by at least 0.8. 
(3) For zoom-in pairs, to avoid trivial cases where the good crop nearly overlaps with the original, we filter out pairs in which the good crop occupies more than 60\% of the original image.

\subsubsection{Composition Assessment Model}
\label{subsubsec:assess_model}

To construct the remaining view-change pairs, a good composition assessment model is necessary. 
As described in \cref{subsubsec:shift_zoomin_data}, the shift and zoom-in pairs are built upon cropping datasets with human-provided annotations. 
However, most in-the-wild images lack such annotations.

\vspace{2pt}\noindent\textbf{Composition assessment dataset}. To train such an assessment model, we primarily use composition scoring datasets, CADB~\cite{cadb} and GAIC~\cite{gaic_v2}, and additionally incorporate composition classification datasets, CADB~\cite{cadb} and KU-PCP~\cite{kupcp}, and the aesthetic assessment dataset AVA~\cite{ava}, given their strong relevance to composition assessment.

\vspace{2pt}\noindent\textbf{Composition assessment model}. Following \cite{qinsight,visualqualityr1}, we adopt Qwen2.5-VL-7B~\cite{qwen2.5_vl} as the base model and train it using the GRPO~\cite{grpo} reinforcement learning algorithm. 
The composition assessment and classification results are presented in \cref{tab:res_assess}. 
Our 7B model outperforms all prior assessment models~\cite{vfn, cpc, autophoto}, Q-Align~\cite{qalign}, and the much larger Qwen2.5-VL-32B~\cite{qwen2.5_vl}. 
One qualitative example is shown in \cref{fig:res_assess}, where our model provides detailed reasoning texts alongside the final accurate composition score.

\begin{figure*}
    \centering
    \includegraphics[width=0.95\linewidth]{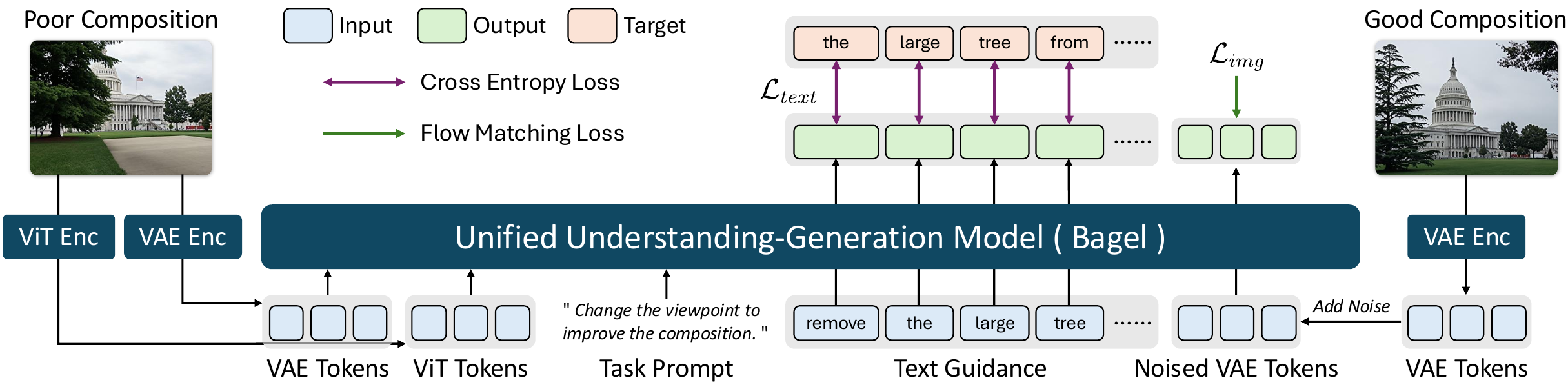}
    \vspace{-10pt}
    \caption{
        \method architecture. We adopt Bagel~\cite{bagel}, a unified understanding–generation model, as the base model. Given a task prompt and a poorly composed image, the model predicts both text guidance and a well-composed example image. 
        The text guidance is optimized using a cross-entropy loss $\mathcal{L}_{text}$ for next-token prediction, while the example image is trained using a flow matching loss $\mathcal{L}_{img}$. 
    }
    \label{fig:model}
    \vspace{-15pt}
\end{figure*}

\subsubsection{View-change Pairs}
\label{subsubsec:view_data}

\vspace{2pt}\noindent\textbf{Pair collection from multi-view data}. A natural source to sample image pairs with varying viewpoints is 3D datasets containing multi-view images. 
DL3DV-10K~\cite{dl3dv} is a large-scale 3D dataset comprising 10K scenes and 51M frames. 
As shown in \cref{fig:data_view}, we evaluate images within each scene using our assessment model, then select up to three best images and ten worst images to form pairs. 
However, even the best frames often lack expert-level composition quality, making it insufficient to rely solely on this dataset.

\vspace{2pt}\noindent\textbf{Pair collection from expert-taken photos}. 
Multi-view data can provide image pairs, but the good images often lack strong composition quality. 
In contrast, expert-taken photos exhibit good composition but lack corresponding poor images to form pairs. 
Thus, as shown in \cref{fig:data_view}, we use the expert-taken photo as good image, and generate a poor view of this image, forming the view-change pair.

\noindent-- \textit{Composition degradation model}. In the first stage, we use the pairs from 3D dataset to train a degradation model (the same to the final model introduced in \cref{sec:model}). 
As depicted in \cref{fig:data_view}a, the model takes a well-composed image along with a degradation instruction such as ``Change the viewpoint to \textit{worsen} the composition'', and generates the corresponding poor-composition image. 
An image reconstruction loss between the predicted poor image and the ground-truth poor image is minimized to optimize the model.

\noindent-- \textit{Degradation on excellent images.} As shown in \cref{fig:data_view}b, in the second stage, we apply the trained degradation model to human-taken photos to synthesize pseudo poor images, forming the final pairs. 
We use two sources of good images: the Unsplash Lite dataset~\cite{unsplash_lite} with 25K professional photographs, and another 10K images taken by ourselves. 
The Unsplash Lite dataset reflects professional-level photography, while our own dataset represents the level of amateur photography, providing complementary data diversity.

\vspace{2pt}\noindent\textbf{Image pair filtering}. We perform three types of filtering to ensure data quality. 
(1) The first discards pairs containing any image with an extreme aspect ratio outside [0.45, 2.2].  
(2) The second aims to ensure the quality of good images. 
(a) File size: we discard images smaller than 200 KB (JPG, 1024 resolution, quality 95), as such images are often overly simplistic (\eg, plain colors or curves). 
(b) Image quality: images with a DeQA-Score~\cite{deqa} below 3.5 are removed. 
(c) Composition: only images with a composition score above 3.0 are retained. 
(d) Art: for the Unsplash Lite dataset~\cite{unsplash_lite}, we filter out images with keywords like ``abstract'', ``painting'', and ``art'', as our focus is on real-world scenes. 
(3) The third aims to enforce view consistency within each image pair. 
VGGT~\cite{vggt} is employed to compute the Field-of-View (FoV) overlap between pairs. Low FoV intersection means poor spatial correspondence and thus leads to exclusion.

\vspace{-6pt}
\section{\method Model}
\label{sec:model}
\vspace{-4pt}

\textbf{Model architecture}. As stated in \cref{subsec:task}, our goal is to predict both a textual instruction and a well-composed example image. 
Bagel~\cite{bagel} is a unified multi-modal model capable of producing both textual and visual outputs, with strong language understanding, visual perception, and image generation capabilities. 
Therefore, we adopt Bagel as our base model for finetuning. 
As illustrated in \cref{fig:model}, given a task prompt and a poorly composed image, the model predicts a textual instruction and a well-composed example image. 
For visual input, Bagel uses two types of vision tokens: VAE tokens encoded by the FLUX VAE~\cite{flux} and ViT tokens extracted by the SigLIP2-so400m/14~\cite{Siglip2} ViT model. 
The VAE tokens, containing pixel-level information, are used for image generation, while the ViT tokens, encoding semantic information, are leveraged for visual understanding. 
The textual instruction is optimized with a cross-entropy loss $\mathcal{L}_{text}$ for next-token prediction, and the example image is optimized with a flow-matching loss $\mathcal{L}_{img}$. 
Note that the generated image attends to the textual instruction through the attention mechanism, enabling instruction-driven example image generation. 
For further architectural details of Bagel, we refer the reader to~\cite{bagel}.

\begin{figure*}
    \centering
    \includegraphics[width=0.95\linewidth]{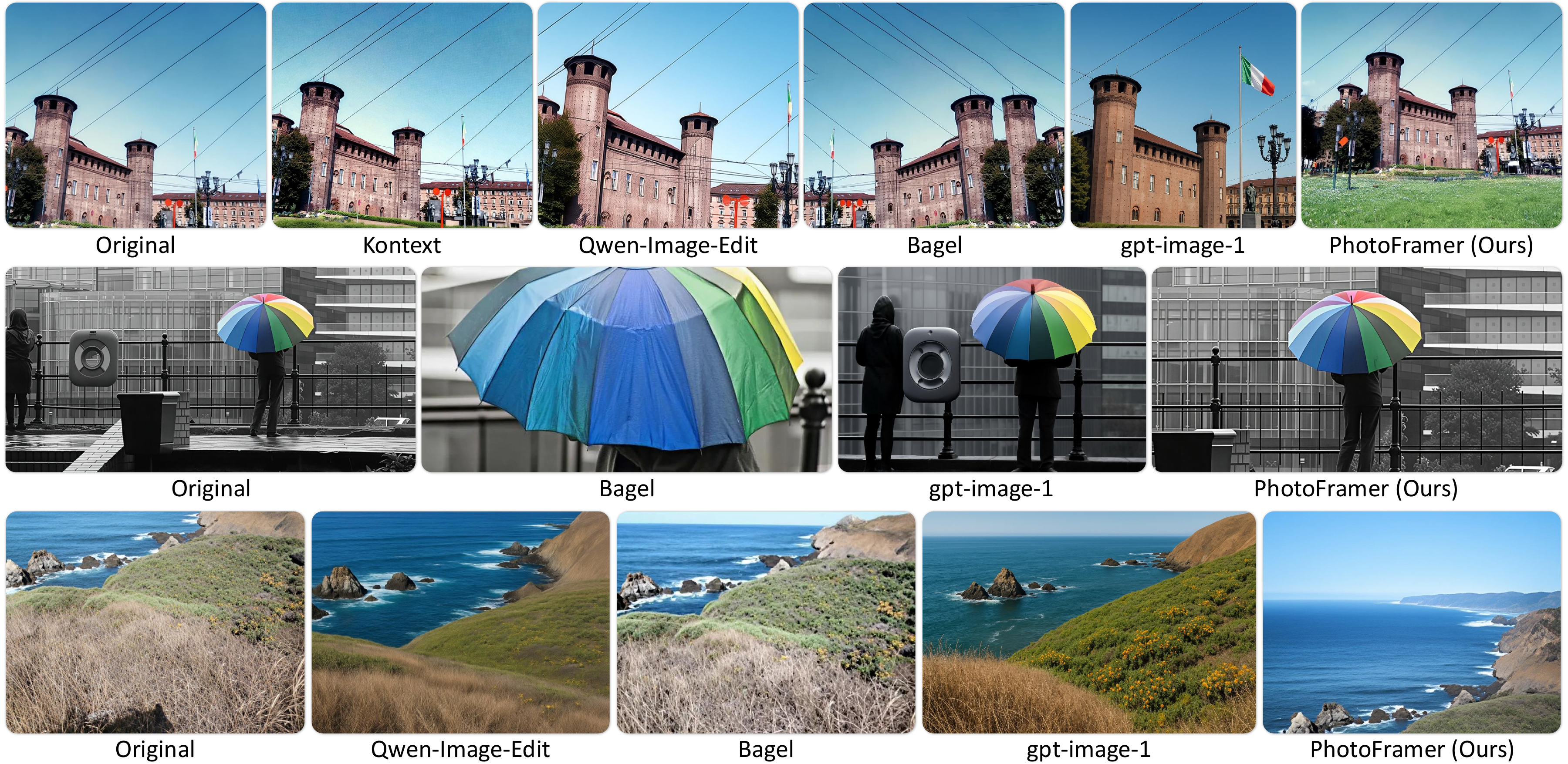}
    \vspace{-10pt}
    \caption{
    Qualitative comparison between our \method\ and baseline methods. Open-source editing models fail to improve composition. The gpt-image-1~\cite{gpt4o} could generate well-composed images but with low fidelity (\ie, it alters semantic details of the original image).
    }
    \label{fig:compare}
    \vspace{-9pt}
\end{figure*}

\newcommand{\iqa}[1]{\textcolor{gray}{#1}}
\begin{table*}[t]
    \centering
    \footnotesize
    \caption{
    Quantitative results of model-generated example images on our benchmark. We compare the model-generated example images with their original images / ground-truth images, evaluated by both GPT-5~\cite{gpt5} and humans. The win rate (\%) is reported as the metric. We also report image quality assessment (with DeQA-Score~\cite{deqa}) and image aesthetic assessment (with Q-Align~\cite{qalign}) results for reference.
    }
    \vspace{-10pt}
    \begin{threeparttable}
    \centering
    \footnotesize
    \setlength\tabcolsep{6pt}
    \begin{tabular}{c|cc|cc|cc|cc}
    \toprule
    \multirow{2}{*}{Method} & \multicolumn{2}{c|}{Shift Task} & \multicolumn{2}{c|}{Zoom-in Task\footnotemark[1]} & \multicolumn{2}{c|}{View-change Task} & \iqa{Quality} & \iqa{Aesthetic} \\
    & GPT-5~\cite{gpt5} & Human & GPT-5~\cite{gpt5} & Human & GPT-5~\cite{gpt5} & Human & \iqa{DeQA~\cite{qalign}} & \iqa{Q-Align~\cite{qalign}} \\
    \midrule
    Kontext~\cite{kontext} & 39.88 / 12.27 & 49.69 / 4.94 & - / 15.52 & - / 5.17 & 46.74 / 15.76 & 48.37 / 5.98  & \iqa{3.88} & \iqa{3.13} \\
    Qwen-Image-Edit~\cite{qwen_image_edit} & 46.01 / 16.56 & 48.43 / 10.49 & - / 39.65 & - / 13.79 & 70.65 / 36.96 & 61.96 / 20.65 & \iqa{4.03} & \iqa{\textbf{3.29}} \\
    Bagel~\cite{bagel} & 27.61 / 14.73 & 38.36 / 8.02 & - / 22.41 & - / 12.07 & 47.28 / 14.13 & 64.13 / 15.22 & \iqa{3.87} & \iqa{3.08} \\
    gpt-image-1~\cite{gpt4o} & 69.93 / 33.99 & 68.46 / 22.37 & - / 52.73 & - / 27.27 & \textbf{84.61} / \textbf{51.65} & 81.52 / 41.30 & \iqa{3.97} & \iqa{3.26} \\
    \midrule
    \method (Ours) & \textbf{80.37} / \textbf{35.58} & \textbf{88.05} / \textbf{43.83} & - / \textbf{67.24} & - / \textbf{48.28} & 82.07 / 50.54 & \textbf{85.87} / \textbf{47.28} & \iqa{\textbf{4.07}} & \iqa{3.17} \\
    \bottomrule
    \end{tabular}
    \end{threeparttable}
    \begin{tablenotes}
    \footnotesize
    \item[1] $^1$ For the zoom-in task, we \textbf{exclude} Zoomed \vs Original because the pair is trivially detectable (\ie, through scale/cropping cues), which induces type preference and inflates win rates. We therefore report only Zoomed \vs Ground-Truth win rate.
    \end{tablenotes}
    \label{tab:main_res}
    \vspace{-15pt}
\end{table*}

\vspace{2pt}\noindent\textbf{Auto prompt design}. As shown in \cref{fig:task}, \method supports three sub-tasks. 
Each task uses predefined prompt templates. 
However, this requires users to explicitly specify the task type, which can be inconvenient. 
To alleviate this issue, we adopt \textit{auto prompts} such that the user does not need to indicate the task. 
We propose two types of auto prompts: 
(1) \textit{Static auto task}, where users may not want to change their viewpoint, thus only shift and zoom-in are allowed. 
(2) \textit{Full auto task}, where all three tasks could be applied. 
For static auto prompts, we use, \eg, ``Refine the composition through shift or zoom-in adjustments''. 
For full auto prompts, we use free-form instructions such as ``Capture this scene with better composition''. 
More examples are provided in \supp. 
During training, task-specific prompts are randomly replaced by auto prompts, enabling the model to self-determine the most suitable operations.

\vspace{2pt}\noindent\textbf{Inference}. The inference process consists of two stages. 
(1) \textit{Understanding}. The model takes visual tokens and a task prompt as inputs, analyzes the current composition, and predicts text guidance describing how to improve it. 
(2) \textit{Generation}. The predicted text guidance is first appended to the inputs. Starting from pure noise tokens, the model progressively denoises to generate latent tokens. After VAE decoding, a well-composed example image is obtained.

\vspace{-6pt}
\section{Experiments}
\label{sec:exp}

\vspace{-3pt}
\subsection{Metrics and Details}
\vspace{-3pt}

\vspace{2pt}\noindent\textbf{Evaluation of example images}. Assessing composition is non-trivial. Although we have trained a composition assessment model, it was used in our dataset construction and thus would bias the evaluation. 
As noted in~\cite{depictqa, pipal}, humans find it easier to compare two images than to rate a single one. 
Therefore, we evaluate each generated example by comparing it with both the original and ground-truth images. 
The comparison is conducted by GPT-5~\cite{gpt5} and humans, and the two win rates are reported as the metric. 
We manually select and carefully examine 200 to 300 samples for each task to construct a benchmark to compute the metric.

\vspace{2pt}\noindent\textbf{Evaluation of text guidance}. 
We evaluate the consistency between the text guidance and the corresponding example images. 
Specifically, we input the original image, the model-improved example image, and the model-predicted text guidance into GPT-5~\cite{gpt5}, and request an evaluation score indicating how accurately the text guidance describes the change from the original image to the example image.

\vspace{2pt}\noindent\textbf{Implementation details}. We follow the setup of Bagel~\cite{bagel} to train our model. 
The VAE encoder and decoder are kept frozen, while the ViT encoder and the main model remain trainable. 
Training is performed on 8 NVIDIA A100 GPUs using AdamW optimizer~\cite{adamw} with a batch size of 8, a learning rate of 2e-5, and 50K training steps. 
An exponential moving average with a decay rate of 0.9999 is applied to stabilize training. 
The two loss terms, $\mathcal{L}_{text}$ and $\mathcal{L}_{img}$, are assigned equal weights. 
All images are resized to a 512 shorter side while preserving the aspect ratio. 
During inference, the number of image generation steps is set to 30.

\vspace{-3pt}
\subsection{Comparison Results}
\vspace{-2pt}

\vspace{2pt}\noindent\textbf{Baselines}. We compare with state-of-the-art editing models, including Kontext~\cite{kontext} and Qwen-Image-Edit~\cite{qwen_image_edit}. 
The original Bagel~\cite{bagel} is also included, along with the powerfully proprietary gpt-image-1~\cite{gpt4o}. 
We adopt Bagel's reasoning-based editing mode, which first generates textual edit instructions and then edits the image accordingly.

\vspace{2pt}\noindent\textbf{Quantitative results of example images} are presented in \cref{tab:main_res}. 
First, open-source editing models fail to improve composition, exhibiting low win rates against both the original and ground-truth images. 
Second, in the view-change task, where the models have greater freedom in generation, their performance is substantially improved. 
Third, gpt-image-1 achieves substantially better results than open-source models, demonstrating strong generalization ability to this new task. 
However, gpt-image-1 often alters semantic details of the original image, as depicted in \cref{fig:compare}. 
Finally, our \method outperforms open-source methods with the highest win rates, and matches or even surpasses gpt-image-1. 
\cref{tab:main_res} shows that our model not only improves composition but also preserves high-level image quality and aesthetics. 
The qualitative examples in \cref{fig:teaser} and \cref{fig:compare} show that \method effectively enhances composition while maintaining high fidelity to the original content.

\begin{table}[t]
    \centering
    \footnotesize
    \setlength\tabcolsep{5.2pt}
    \vspace{-1pt}
    \caption{Quantitative results of consistency between the model-predicted text guidance and corresponding example images.}
    \vspace{-10pt}
    \begin{tabular}{c|ccc|c}
    \toprule
    Task & Shift & Zoom-in & View-change & Average \\
    \midrule
    Bagel~\cite{bagel} & 77.01 & 84.82 & 87.47 & 83.10 \\
    \method (Ours) & \textbf{91.96} & \textbf{92.59} & \textbf{91.52} & \textbf{92.02} \\
    \bottomrule
    \end{tabular}
    \label{tab:text_img_align}
    \vspace{-17pt}
\end{table}

\vspace{2pt}\noindent\textbf{Evaluation results of text guidance} are provided in \cref{tab:text_img_align}. 
The original Bagel model has exhibited reasonable consistency between text guidance and the corresponding example images across the three composition instruction sub-tasks. 
After finetuning on our constructed datasets, this consistency is further and stably improved (92.02\% \vs 83.10\%).

\vspace{-3pt}
\subsection{Ablation Studies and Discussions}
\vspace{-2pt}

\begin{figure}
    \centering
    \includegraphics[width=1.0\linewidth]{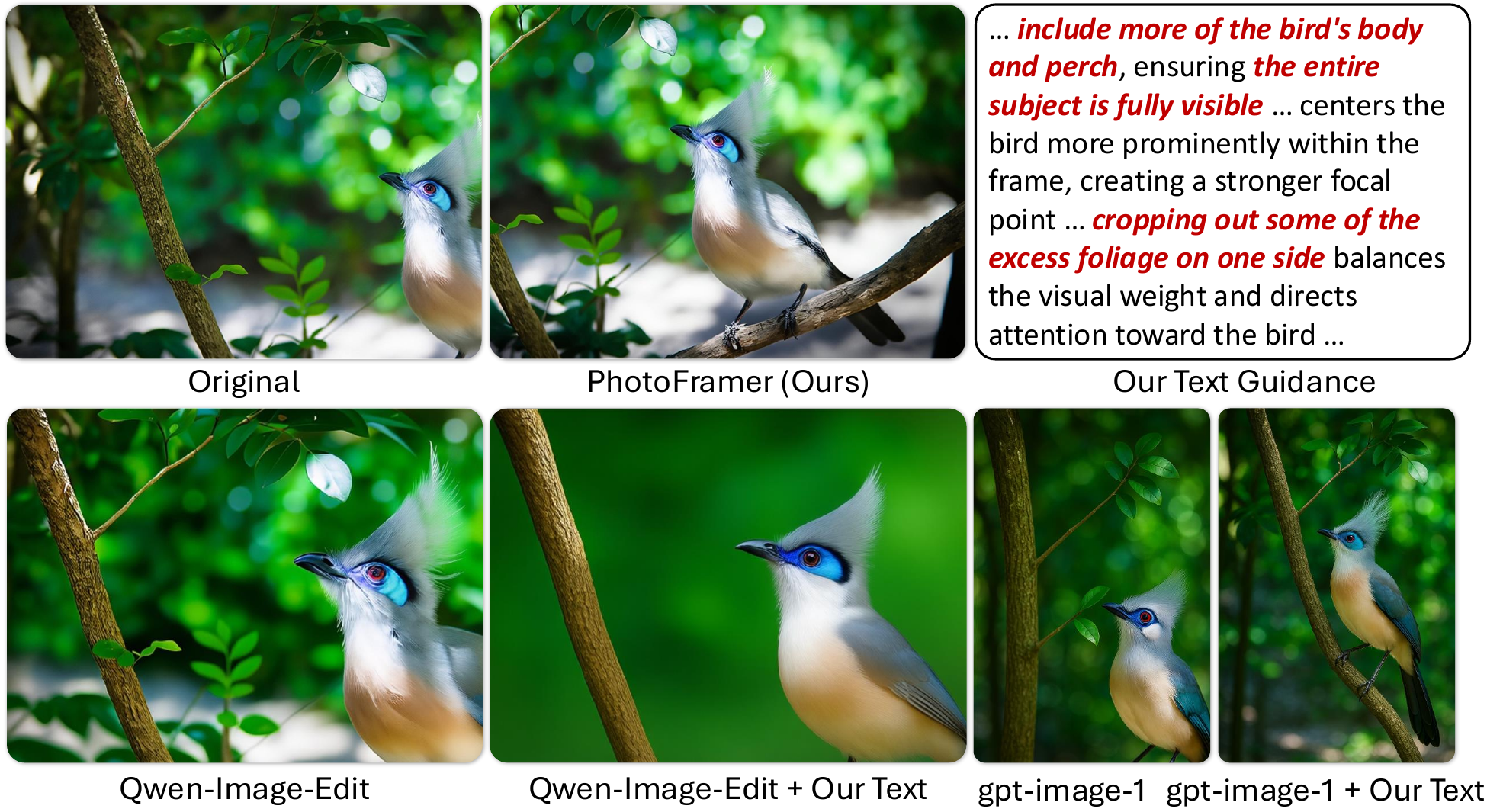}
    \vspace{-18pt}
    \captionof{figure}{Our generated text guidance cannot be directly utilized by Qwen-Image-Edit, even though it employs an LLM as the text encoder. In contrast, gpt-image-1 benefits from text guidance (\ie, including the entire body of the bird), albeit with lower fidelity.}
    \label{fig:base_use_our_text}
    \vfill
    \vspace{2pt}
    \centering
    \includegraphics[width=1.0\linewidth]{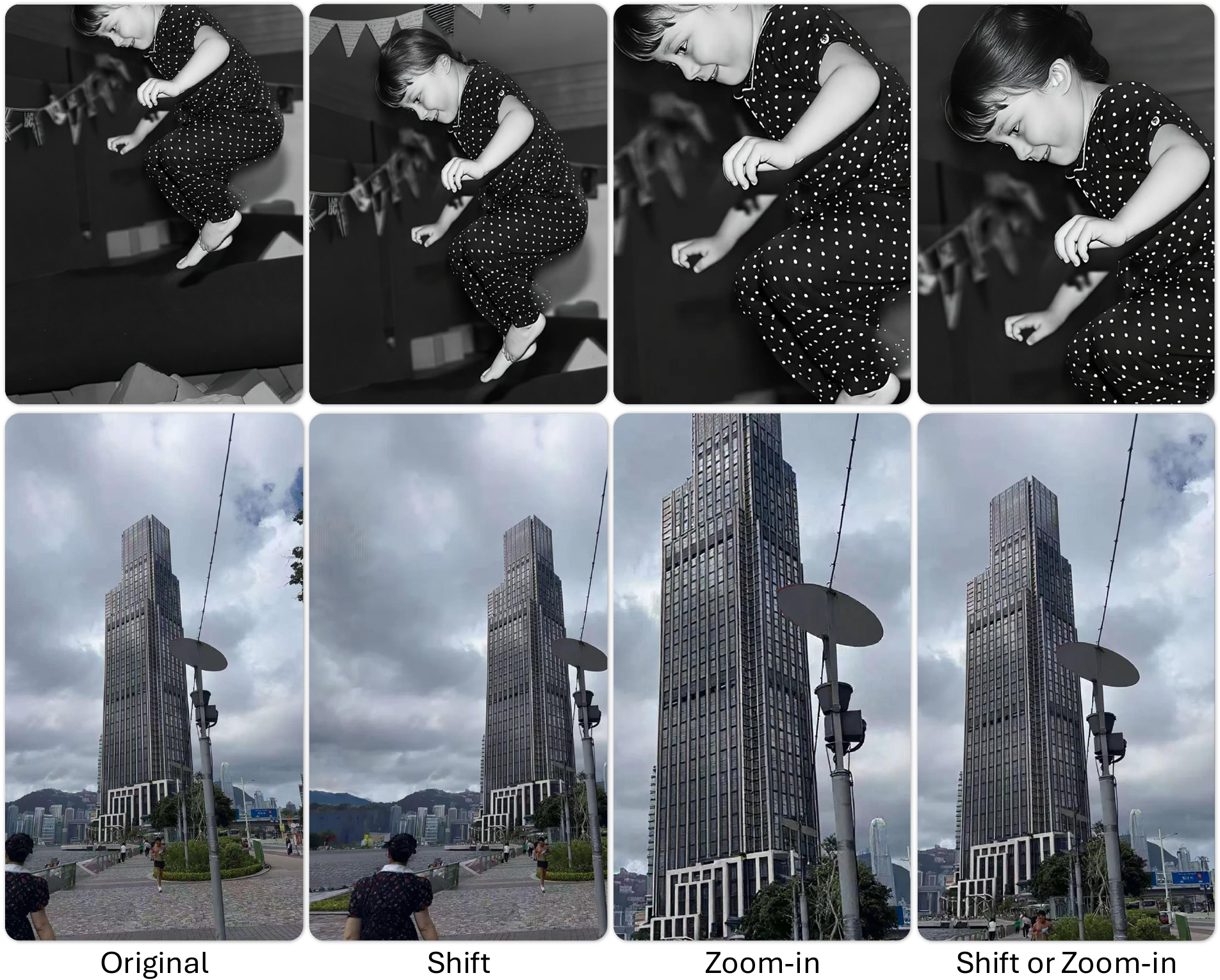}
    \vspace{-18pt}
    \captionof{figure}{Illustration of auto prompt. Given an auto prompt ``refine the composition using shift or zoom-in’’, our model does not merely select one task type but adaptively fuses multiple operations to produce a better composition. Text guidance is omitted.}
    \label{fig:auto_task}
    \vspace{-15pt}
\end{figure}

\textbf{Text guidance alone is not enough}. 
A natural question arises for the unified framework: can we rely solely on the text guidance and feed it into other editing models to generate example images? 
As shown in \cref{fig:base_use_our_text}, our generated text guidance cannot be directly utilized by Qwen-Image-Edit, though it employs a Large Language Model (LLM) as the text encoder. 
Moreover, directly using our text guidance even degrades the fidelity of example images. 
In contrast, when provided with our text guidance (\ie, ``include more of the bird's body and perch’’), gpt-image-1 successfully includes the entire bird, showing strong instruction-following ability. 
However, its fidelity remains unsatisfactory.

\begin{figure}
    \centering
    \includegraphics[width=1.0\linewidth]{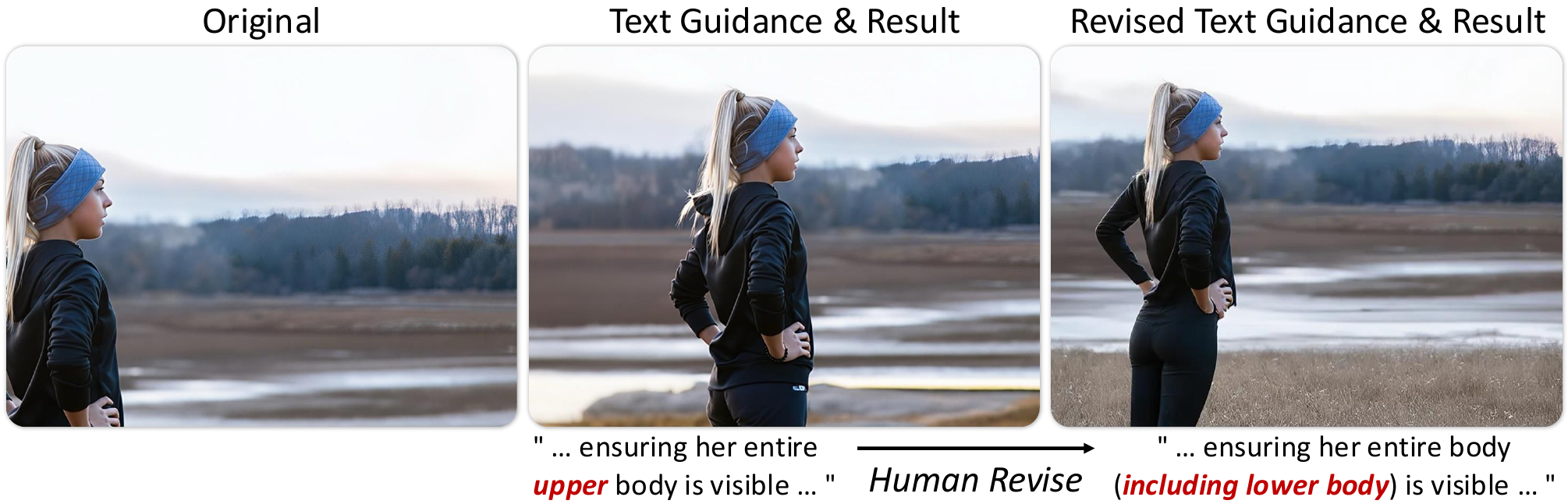}
    \vspace{-19pt}
    \captionof{figure}{
    Text guidance is important for image generation. If we manually revise a few key words (\ie, remove ``upper'', add ``including lower body''), the generated image will be quite different.
    }
    \label{fig:change_text}
    \vspace{2pt}
    \centering
    \includegraphics[width=1.0\linewidth]{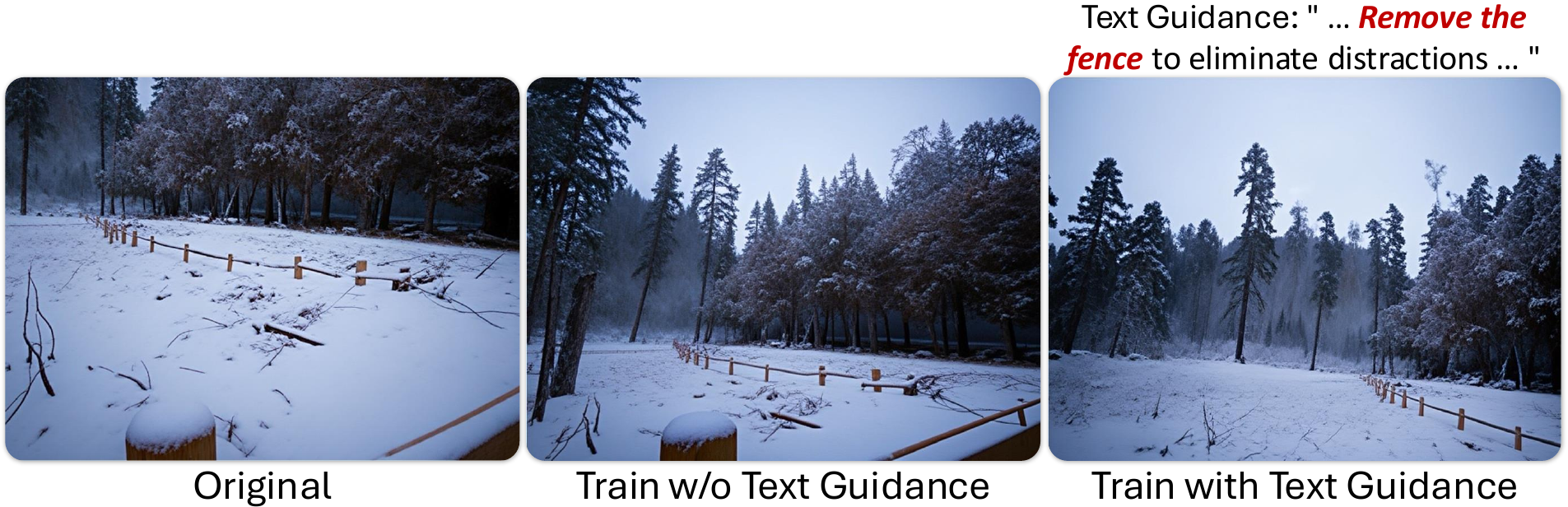}
    \vspace{-19pt}
    \captionof{figure}{Training without text guidance fails to remove the fence distractions, although it successfully includes the sky, whereas training with text guidance predicts to ``remove the fence'' in textual output and generates a well-composed image without fence.}
    \label{fig:text_vs_notext}
    \vspace{2pt}
    \centering
    \includegraphics[width=1.0\linewidth]{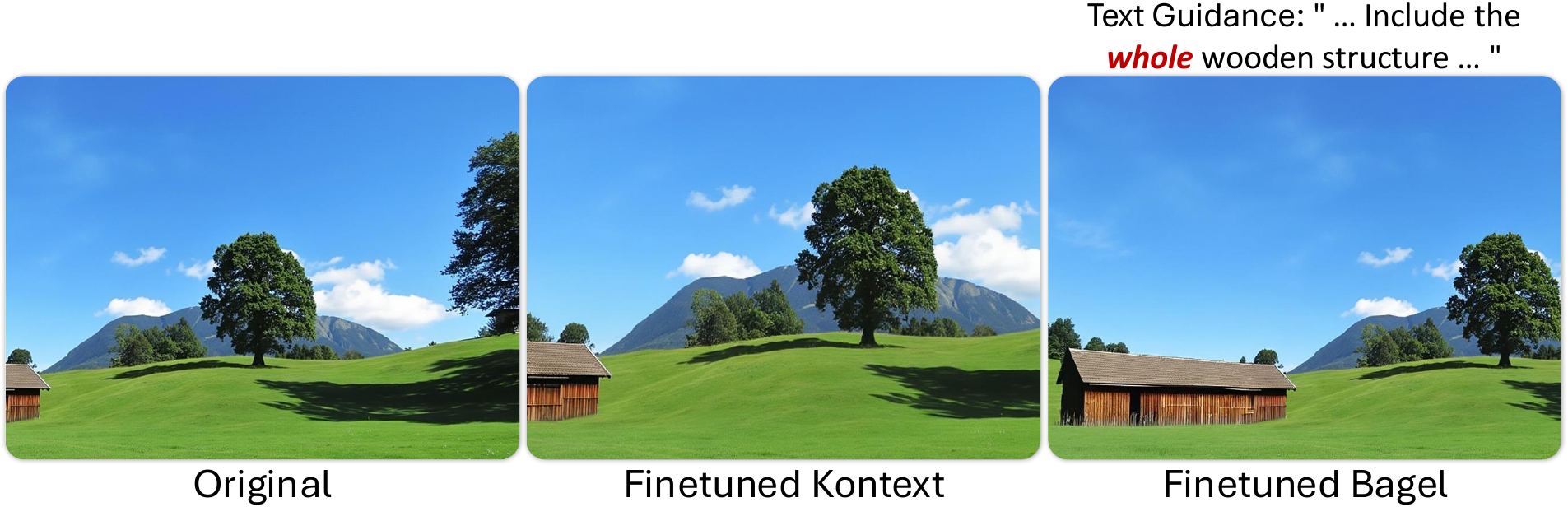}
    \vspace{-19pt}
    \caption{Finetuned Bagel (\ie, on both textual and visual data) successfully includes the \textit{whole} wooden house as in the text guidance, outperforming finetuned Kontext (\ie, on visual data only).}
    \label{fig:bagel_vs_kontext}
    \vspace{-12pt}
\end{figure}

\vspace{2pt}\noindent\textbf{Example image alone is not enough}. First, as depicted in \cref{fig:change_text}, textual guidance plays a crucial role in generating example images. 
Even revising a few key words could lead to dramatically different results. 
Second, we train Bagel using only image pairs without text guidance. As illustrated in \cref{fig:text_vs_notext}, without textual input, the model fails to remove the foreground fence, although it successfully includes the sky. 
In contrast, when trained with text guidance, the model explicitly learns to ``remove the fence'' and successfully follows this instruction to generate a better example image. 
Third, we fine-tune Kontext (which does not support textual training) using our collected image pairs. 
As shown in \cref{fig:bagel_vs_kontext}, Kontext can partially shift the wooden house toward the center but fails to include it entirely. 
In contrast, the fine-tuned Bagel predicts to ``include the \textit{whole} wooden structure'' and generates a well-composed image accordingly.

\vspace{2pt}\noindent\textbf{Task prompt}. As illustrated at the top of \cref{fig:auto_task}, different task prompts lead \method\ to apply different operations to improve composition. 
Notably, as shown at the bottom of \cref{fig:auto_task}, when specific tasks fail, the auto task can still produce strong results. 
The original image was randomly taken in a city and contains a distraction (\ie, the back of a half-visible woman). 
The shift task mistakenly treats this distraction as the main subject and attempts to center it, while the zoom-in task crops too tightly, cutting off the top of the building. 
With the auto prompt, where the model can adaptively apply shift and/or zoom-in, \method\ effectively fuses multiple adjustments to achieve a better composition.

\vspace{-2pt}
\section{Conclusions}
\label{sec:conclusions}
\vspace{-2pt}

We introduce \method, a multi-modal composition instruction model built upon a hierarchical task paradigm, curated datasets, and a unified understanding–generation framework, to guide photographic composition through actionable textual instructions and example image generation.

\noindent\textbf{Acknowledgment}. 
This research work was supported by National Natural Science Foundation of China (Grant No. 62276251), RGC Early Career Scheme (ECS) No. 24209224, and the Joint Lab of CAS-HK. 

{
    \small
    \bibliographystyle{ieeenat_fullname}
    \bibliography{main}
}
\clearpage
\renewcommand\thefigure{A\arabic{figure}}
\renewcommand\thetable{A\arabic{table}}  
\renewcommand\theequation{A\arabic{equation}}
\setcounter{equation}{0}
\setcounter{table}{0}
\setcounter{figure}{0}
\appendix

\section*{Appendix}
\section{More Results}

\definecolor{myblue}{rgb}{0.18, 0.43, 0.73}
\definecolor{myred}{rgb}{0.75, 0., 0.}

We have added more qualitative results in \cref{supp:fig:res_shift,supp:fig:res_zoomin,supp:fig:res_view}, including the poorly composed input, the predicted text guidance, and the model-generated well-composed image, for all three sub-tasks.
In the predicted text guidance, actionable suggestions are highlighted in \textbf{\textcolor{myblue}{blue}}, while hallucinated descriptions are marked in \textbf{\textcolor{myred}{red}}.
Overall, our \method could provide clear and practically useful composition suggestions that correspond well to the generated example well-composed images.
However, in some cases, small hallucinations still occur. For instance, in the shift task, the model may confuse left and right directions, leading to incorrect spatial descriptions.

\begin{figure}[h!]
    \centering
    \vspace{-5pt}
    \includegraphics[width=1.0\linewidth]{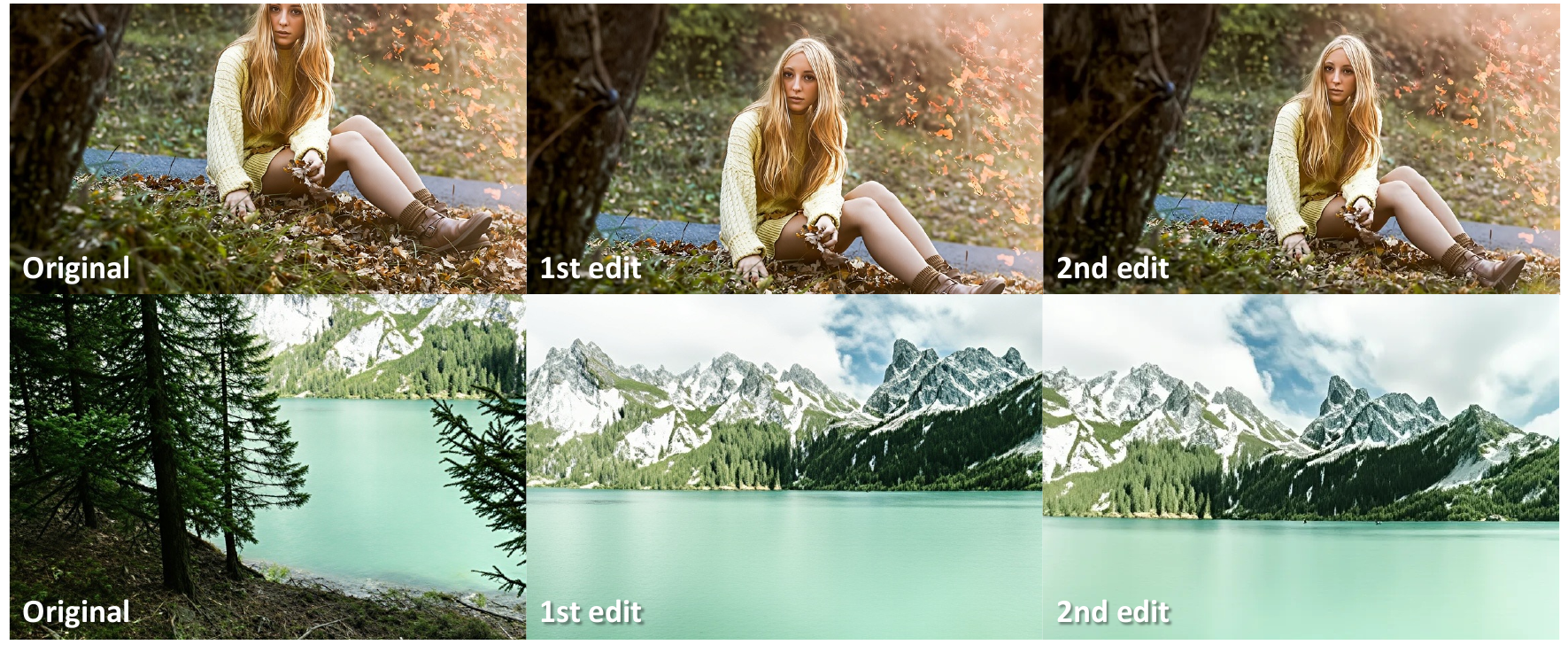}
    \vspace{-15pt}
    \caption{Our \method can be applied to refine the image composition iteratively.}
    \label{supp:fig:iterative}
    \vspace{-10pt}
\end{figure}

We also test if one iteration of view improvement would be sufficient to get the desired result or whether iterative approach is needed.
In many cases, a single round of guidance is enough to yield a clear improvement.
Furthermore, as illustrated in \cref{supp:fig:iterative}, our approach can be applied to refine the image composition iteratively.

\section{More Details of Dataset Construction}

In this section, we provide more details of our dataset construction, including the optimization to obtain composition scores for CPC dataset (\cref{supp:subsec:score_cpc}), composition assessment model training (\cref{supp:subsec:comp_assess}), details of view-change pairs collection (\cref{supp:sub:view_detail}), more dataset statistics (\cref{supp:subsec:statis}), and task prompt details (\cref{supp:subsec:task_prompt}).

\subsection{Score Optimization of CPC Dataset}
\label{supp:subsec:score_cpc}

As stated in Sec.~\textcolor{red}{3.2.1} of the main paper, 
the composition scores are required for each crop to construct shift or zoom-in pairs. 
However, the CPC dataset~\cite{cpc} contains only selected good and best crops, which cannot be directly used to construct the pairs. 
Therefore, we build a mathematical model to infer scores for all crops.

\vspace{2pt}\noindent\textbf{Annotation of CPC dataset}. Given an original image and its $N$ crops ($N{=}24$ in CPC), one annotator first selects 8$\sim$20 good crops, and then chooses the 3 best crops from these good ones. 
Each image is annotated by 6 annotators. 
By averaging the annotations from the 6 annotators, we obtain the probability that crop $i$ is good, $p^{\text{good}}_i$, and the probability that it belongs to the top-3 set, $p^{\text{top3}}_i$.

\vspace{2pt}\noindent\textbf{Assumption}. Annotators are required to select 8$\sim$20 good crops out of all 24 crops, where good crops constitute a relatively large proportion. 
Therefore, we assume that a crop $i$ is considered good if its score $s_i$ exceeds a threshold, and this decision \textit{depends only on its own quality without pairwise comparison}. 
In contrast, selecting the top-3 crops requires careful comparison among candidates.
Thus, the decision necessarily \textit{depends on relative scores between crops}.

\vspace{2pt}\noindent\textbf{Mathematical modeling}. According to the above assumption, whether a crop is considered good or not \textit{depends only on its own quality without pairwise comparison}. Therefore, we directly model the probability of a crop being good as a sigmoid function of its own score: 
\begin{equation}
    \hat{p}^{\text{good}}_i = \sigma(s_i),
\end{equation}
where $\sigma(\cdot)$ denotes the sigmoid activation function.

For the top-3 crops, careful comparison among crop candidates must be required. Therefore, we first model the probability of each crop being selected as the top-1 using a softmax (\ie, comparison) over all crop scores: 
\begin{equation}
    \hat{p}^{\text{top1}} = {\rm softmax}(\alpha [s_0, \dots, s_i, \dots, s_N]) \in \mathbb{R}^{N},
\end{equation}
where $\alpha$ is a scaling factor, and it is set to 2. 
We can view this problem as a sampling problem, where $p^{\text{top1}}_i$ denotes the probability of sampling crop $i$. 
Selecting the best top-3 can then be formulated as sampling without replacement. 
The probability that crop $i$ is selected into the top-3 is:
\begin{equation}
\begin{aligned}
\hat{p}^{\text{top3}}_i ={} &
p^{\text{top1}}_i \\
& \qquad\qquad (\rightarrow \quad \text{crop $i$ is chosen first}) \\
{}+&
p^{\text{top1}}_i \sum_{j \ne i} \frac{p^{\text{top1}}_j}{1 - p^{\text{top1}}_j} \\
& \qquad\qquad (\rightarrow \quad \text{crop $i$ is chosen second}) \\
{}+&
p^{\text{top1}}_i \sum_{j \ne i} \sum_{\substack{k \ne i \\ k \ne j}}
        \frac{p^{\text{top1}}_j p^{\text{top1}}_k}{(1 - p^{\text{top1}}_j)\,\big(1 - p^{\text{top1}}_j - p^{\text{top1}}_k\big)} \\
& \qquad\qquad (\rightarrow \quad \text{crop $i$ is chosen third})
\end{aligned}
\label{supp:eq:no_replace}
\end{equation}
This formulation is difficult to compute and optimize.
Therefore, considering that there are a total of 24 crops for each image and $24 \gg 3$, we approximate the process of ``sampling the top-3 without replacement'' as ``3 independent samplings with replacement''. 
\begin{equation}
    \hat{p}^{\text{top3}}_i = 1 - (1 - p^{\text{top1}}_i)^3.
\label{supp:eq:replace}
\end{equation}
Mathematically, \cref{supp:eq:no_replace} can be approximated by \cref{supp:eq:replace} when $N$ is large and $\max(p^{\text{top1}}_i)$ is relatively small. 
First, in CPC, each image contains 24 crops, and $N{=}24$ is sufficiently larger than the sampling size of 3. 
Second, based on the human-annotated $p^{\text{top3}}_i$, we observe that in most cases no single crop is selected among the top-3 by more than 4 out of 6 annotators, indicating that multiple crops are competitive. 
Thus, no crop clearly dominates others, \ie, $\max(p^{\text{top1}}_i)$ remains small, supporting that \cref{supp:eq:no_replace} can be well approximated by \cref{supp:eq:replace}.

\vspace{2pt}\noindent\textbf{Loss function and optimization}. 
We treat the composition scores $s_i$ as learnable parameters, from which we estimate the probability that crop $i$ is good, $\hat{p}^{\text{good}}_i$, and the probability that it belongs to the top-3 set, $\hat{p}^{\text{top3}}_i$. 
We define the loss as the discrepancy between the estimated probabilities and human-annotated ground truth:
\begin{equation}
    \mathcal{L} = \sum_{i=1}^{N} \left((\hat{p}^{\text{good}}_i - p^{\text{good}}_i)^2 
    + \beta \, (\hat{p}^{\text{top3}}_i - p^{\text{top3}}_i)^2 \right),
\end{equation}
where $\beta$ is a weighting factor, and it is set to 2 by default to emphasize the top-3 term.

We optimize the scores for each original image in the CPC dataset independently. 
We first apply the L-BFGS~\cite{lbfgs} optimizer (lr=1.0, max\_epochs=10, max\_iter=200, history\_size=10), which is efficient (less than 1 second per image). 
However, it can be unstable, leading to optimization failures. 
Therefore, when the final loss $\mathcal{L} > 0.5$, we switch to the Adam~\cite{adam} optimizer (lr=2e-3, max\_epochs=5000), which is slower (20$\sim$30 seconds per image) but more stable. 
Overall, the average loss converges to around 0.36. 
The mean absolute errors are 0.0602 between $\hat{p}^{\text{good}}_i$ and $p^{\text{good}}_i$, and 0.0426 between $\hat{p}^{\text{top3}}_i$ and $p^{\text{top3}}_i$, indicating sufficiently accurate probability estimation for practical use.

\vspace{2pt}\noindent\textbf{Post processing}. The optimized composition scores are clipped and normalized to the range [1,5] for further use.

\subsection{Details of Composition Assessment Model}
\label{supp:subsec:comp_assess}

\textbf{Composition assessment dataset collection}. 
To train the composition assessment model, we primarily use composition scoring datasets, CADB~\cite{cadb} and GAIC~\cite{gaic_v2}, where each image is annotated with a composition score. 
We additionally incorporate composition classification datasets, CADB~\cite{cadb} and KU-PCP~\cite{kupcp}, and the aesthetic assessment dataset AVA~\cite{ava}, due to their strong relevance to the composition evaluation. 
The composition scores or aesthetic scores of each dataset are normalized to the range of [1, 5]. 
There are 13 composition categories in CADB dataset, \ie, [center, curved, diagonal, fill the frame, golden ratio, horizontal, pattern, radial, rule of thirds, symmetric, triangle, vanishing point, vertical].
KU-PCP dataset contains 9 composition classes including [center, curved, diagonal, horizontal, pattern, rule of thirds, symmetric, triangle, vertical].

We perform data resampling strategy on the AVA and GAIC datasets to mitigate data imbalance and redundancy. 
First, the AVA dataset contains an excessive number of mid-quality images (\ie, 97.35\% in the [2,4) score range), while low-quality (\ie, [1,2)) and high-quality (\ie, [4,5]) samples account for only 0.87\% and 1.78\%, respectively, which may bias model training. 
Therefore, we re-sample AVA to increase the proportions of low- and high-quality images, as illustrated in \cref{supp:tab:ava_resample}. 
Second, the GAIC dataset includes 288K cropped images with MOS labels, but these are derived from only 3,336 original images, each contributing over 80 crops, resulting in insufficient data diversity. 
To address this, we sample 20\% of GAIC, ensuring a diverse MOS distribution, which yields approximately 56K samples for training and evaluation. 
The final dataset statistics are summarized in \cref{supp:tab:comp_data_statis}.

\begin{table}[t]
    \centering
    \scriptsize
    \setlength\tabcolsep{2.9pt}
    \caption{Re-sampling AVA dataset to increase the proportion of low (\ie, [1,2)) and high (\ie, [4,5]) score range.}
    \vspace{-5pt}
    \begin{tabular}{c|cccc}
    \toprule
    Score range & [1,2) & [2,3) & [3,4) & [4,5] \\
    \midrule
    Original & 2,051 / 0.87\% & 91,401 / 38.80\% & 137,946 / 58.55\% & 4,200 / 1.78\% \\
    Re-sampled & 2,051 / 4.75\% & 17,408 / 40.28\% & 19,562 / 45.26\% & 4,200 / 9.72\% \\
    \bottomrule
    \end{tabular}
    \label{supp:tab:ava_resample}
\vfill
\vspace{8pt}
    \centering
    \footnotesize
    \setlength\tabcolsep{8.5pt}
    \caption{Statistics of our collected and re-sampled datasets to train the composition assessment model.}
    \vspace{-5pt}
    \begin{tabular}{c|ccc|cc}
    \toprule
    & \multicolumn{3}{c|}{Assessment} & \multicolumn{2}{c}{Classification} \\
    & CADB & GAIC & AVA & CADB & KU-PCP \\
    \midrule
    Train & 8,547 & 48,140 & 43,221 & 8,547 & 4,244 \\
    Test & 950 & 8,472 & 4,557 & 950 & - \\
    \bottomrule
    \end{tabular}
    \label{supp:tab:comp_data_statis}
    \vspace{-10pt}
\end{table}

\vspace{2pt}\noindent\textbf{Composition assessment model training using GRPO}. 
Following Q-Insight~\cite{qinsight} and VisualQuality-R1~\cite{visualqualityr1}, we adopt Qwen2.5-VL-7B~\cite{qwen2.5_vl} as the base model and train it using the GRPO~\cite{grpo} reinforcement learning algorithm. 
Specifically, for each question, we request the model to first output the thinking process in \texttt{<think></think>} tags and then output the final answer in \texttt{<answer><answer>} tags.
We repeatedly request the model by $N$ times to obtain $N$ outputs, \{$o_1$, $o_2$, ..., $o_N$\}. 
Then, for each output, we extract the answer between \texttt{<answer><answer>} tags, and calculate rewards \{$r_1$, $r_2$, ..., $r_N$\} by comparing the answers and ground truth. 
By calculating the mean and standard deviation of the rewards, the relative advantages of each response can be obtained as follows. 
\begin{equation}
    \hat{A}_i = 
    \frac{r_i - \mathrm{mean}\big(\{r_1, r_2, ..., r_N\}\}\big)}
    {\mathrm{std}\big(\{r_1, r_2, ..., r_N\}\big)},
\end{equation}
where $\hat{A}_i$ denotes the normalized relative advantage (quality) of the $i$-th response. 
Overall, GRPO guides the policy model to prioritize higher-quality responses that receive higher reward values within each group. 
After obtaining $\hat{A}_i$, GRPO computes the ratio between the probabilities of the same response under the updated policy $\pi_{\theta_{\text{new}}}$ and the previous policy $\pi_{\theta_{\text{old}}}$, denoted as $\rho_i$. 
To prevent excessively large policy updates and stabilize training, $\rho_i$ is constrained within the range $[1-\delta, 1+\delta]$. 
In addition, to maintain proximity to the reference distribution $\pi_{\text{ref}}$, a KL-divergence penalty weighted by $\beta$ is introduced. 
Finally, the optimization objective of GRPO can be formulated as follows:
\begin{equation}
\begin{aligned}
\mathcal{J}(\theta)
&= \mathbb{E}_{q \sim \mathcal{Q},o_i\sim \pi_{\theta_{\text{old}}}}\!\{ \\
&\left[
\min\!\left(
\rho_i \hat{A}_i,
\operatorname{clip}\!\left(\rho_i,1-\delta,1+\delta\right)\hat{A}_i \right)\right] \\
&-\beta\,\mathbb{D}_{\mathrm{KL}}\!\left(\pi_{\theta_{\text{new}}}\,\|\,\pi_{\text{ref}}\right)
\},
\end{aligned}
\label{supp:eq:loss_grpo}
\end{equation}
where $\rho_i = \pi_{\theta_{\text{new}}}(o_i \mid q)\,/\,\pi_{\theta_{\text{old}}}(o_i \mid q)$, 
$\mathcal{Q}$ denotes the set of candidate questions, and $\mathbb{D}_{\mathrm{KL}}$ is the KL regularization term. 
The reference policy $\pi_{\text{ref}}$ is typically a frozen pre-trained MLLM. 
Overall, GRPO balances consistent policy updates with strong reward signals, enabling stable yet effective optimization.

\begin{figure*}[t]
    \centering
    \includegraphics[width=1.0\linewidth]{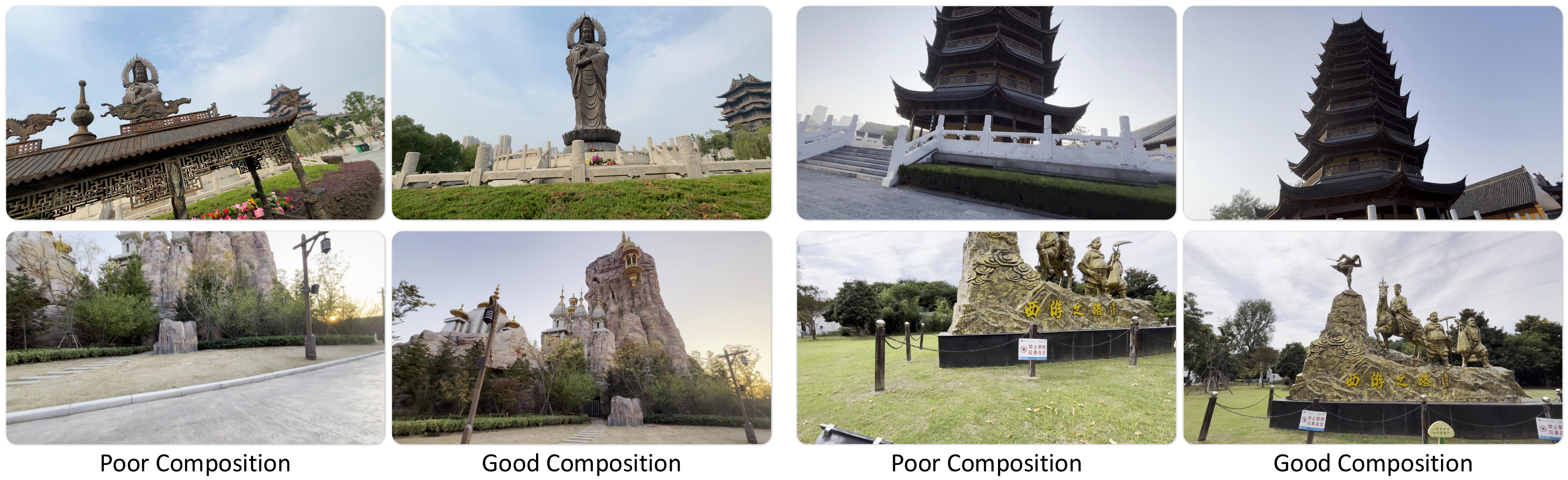}
    \vspace{-15pt}
    \caption{View-change pairs sampled by our composition assessment model from the DL3DV dataset.}
    \label{supp:fig:dl3dv_good_poor}
    \vspace{-10pt}
\end{figure*}

\vspace{2pt}\noindent\textbf{Reward calculation}. To optimize \cref{supp:eq:loss_grpo}, we need to calculate reward $r_i$ for each output $o_i$.

\noindent-- \textit{Format reward} evaluates whether the reasoning steps are properly enclosed within the \texttt{<think></think>} tags, and whether the final answer is correctly enclosed within the \texttt{<answer></answer>} tags. 
The format reward $r_i^{\text{format}}$ is set to 1 if the $i$-th response satisfies above conditions; otherwise, it is set to 0.

\noindent-- \textit{Score reward} evaluates whether the predicted composition or aesthetic score is accurate. 
We first extract the model-predicted score within the \texttt{<answer></answer>} tags, then compute its error with respect to the ground-truth score. 
If the error is equal to or smaller than a threshold $\sigma$, the score reward $r_i^{\text{score}}$ for the $i$-th response is set to 1; otherwise, it is set to 0. The threshold $\sigma$ is set as 0.4 empirically.

\noindent-- \textit{Classification reward} evaluates whether the predicted composition categories match the ground-truth categories. 
We provide the model with all possible composition categories and ask it to select one to three of the most suitable ones (since the ground-truth annotations may contain one to three types). 
Let the ground truth contain $n$ composition types, the model predict $m$ types, and let $k$ be the number of correctly predicted types. The classification reward is then computed as $r_i^{\text{class}} = k / \max(m, n)$.

Finally, the overall reward for the composition or aesthetic assessment task is defined as: 
\begin{equation}
    r_i =
    \begin{cases}
        r_i^{\text{format}} + r_i^{\text{score}}, & \text{if } r_i^{\text{format}} = 1,\\[4pt]
        0, & \text{otherwise},
    \end{cases}
\end{equation}
and the overall reward for composition classification task is: 
\begin{equation}
    r_i =
    \begin{cases}
        r_i^{\text{format}} + r_i^{\text{class}}, & \text{if } r_i^{\text{format}} = 1,\\[4pt]
        0, & \text{otherwise}.
    \end{cases}
\end{equation}
With the final reward $r_i$ for each output $o_i$, we then optimize \cref{supp:eq:loss_grpo} to train the composition assessment model.

\vspace{2pt}\noindent\textbf{Implementation details}. We adopt Qwen2.5-VL-7B~\cite{qwen2.5_vl} as the base model, with all model components, including the vision encoder, vision–text connector, and large language model, kept trainable. 
To ensure the data balance, the CADB and KU-PCP datasets are duplicated by 5 times. 
For each question, the number of generated responses $N$ is set to 8. 
Training is performed for 1 epoch with a batch size of 32 on 8 NVIDIA A6000 GPUs using 4 gradient-accumulation steps. 
We employ the AdamW optimizer~\cite{adamw} with a learning rate of 1e-6. 
The coefficients $\beta$ and $\epsilon$ in \cref{supp:eq:loss_grpo} are set to 0.04 and 0.2, respectively.

\begin{figure*}[t]
    \centering
    \includegraphics[width=1.0\linewidth]{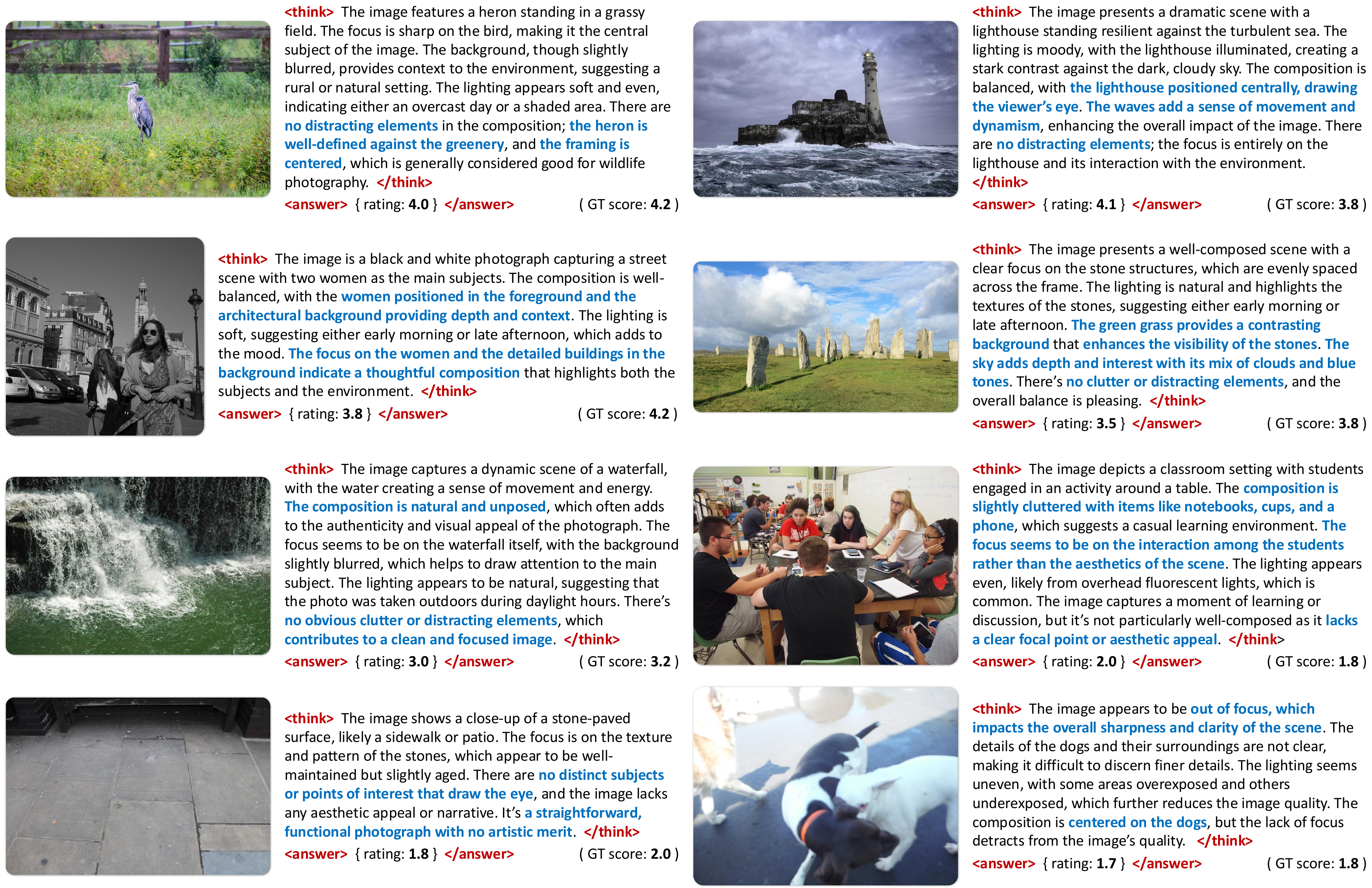}
    \vspace{-15pt}
    \caption{Qualitative results of image composition assessment on CADB dataset.}
    \label{supp:fig:res_assess}
\vfill
\vspace{20pt}
    \centering
    \includegraphics[width=1.0\linewidth]{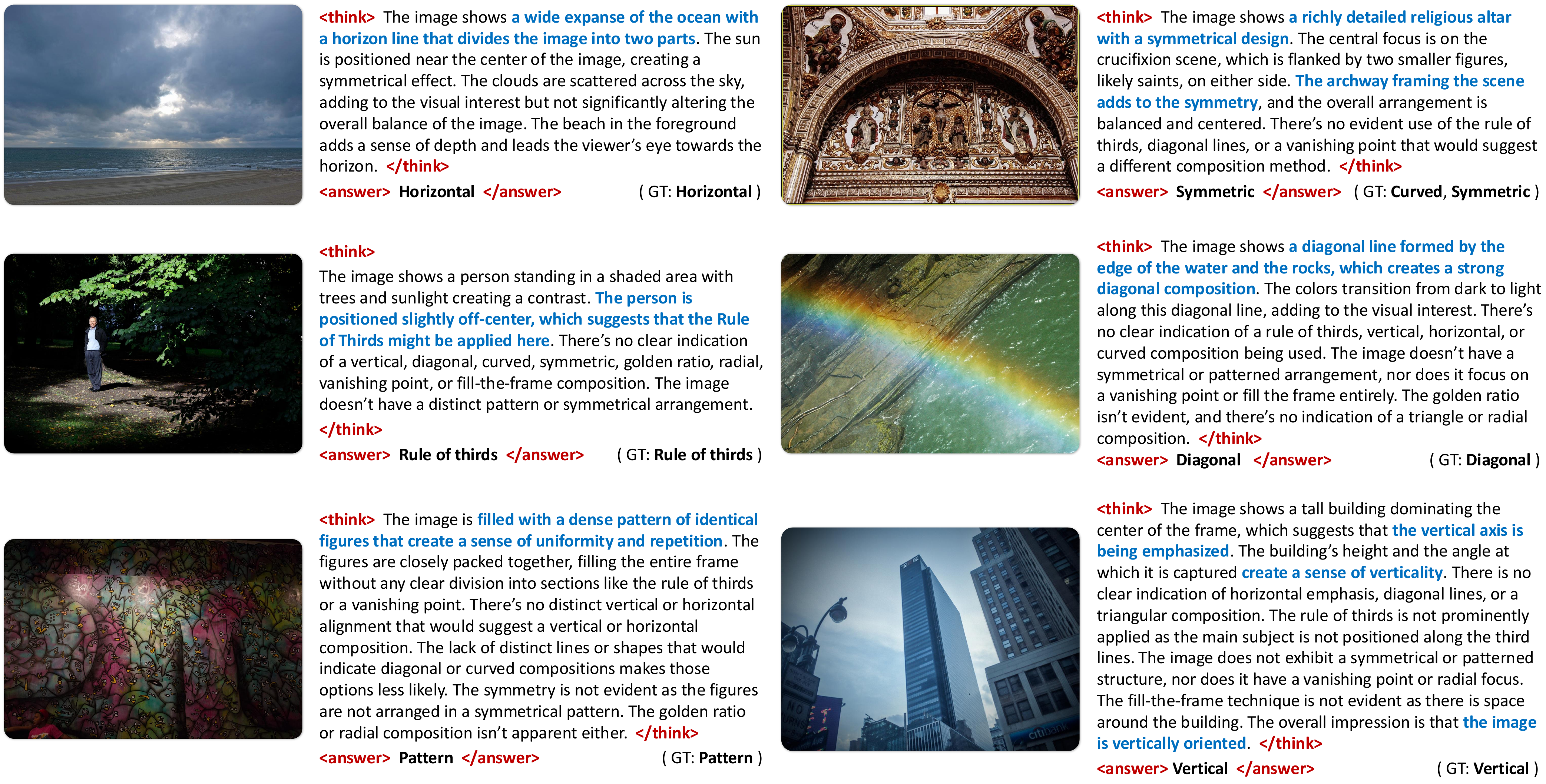}
    \vspace{-15pt}
    \caption{Qualitative results of image composition classification on CADB dataset.}
    \label{supp:fig:res_class}
\end{figure*}

\vspace{2pt}\noindent\textbf{Composition assessment results}. Quantitative results of our composition assessment model have been reported 
in Tab.~\textcolor{red}{2} of the main paper. 
Here, we provide additional qualitative examples in \cref{supp:fig:res_assess} and \cref{supp:fig:res_class}. 
The results show that our model accurately understands compositional structure and produces reliable composition scores accompanied by detailed and coherent reasoning.

\subsection{Details of View-change Pairs Collection}
\label{supp:sub:view_detail}

\textbf{View-change pairs sampled from the multi-view 3D dataset}. 
We visualize the sampled View-change pairs from the DL3DV~\cite{dl3dv} dataset in \cref{supp:fig:dl3dv_good_poor}. 
As shown, even the good-composition images in DL3DV reflect ordinary, casually captured viewpoints rather than expert-level compositions. 
This motivates the introduction of the degradation model, which is then applied to expert-taken photos to further construct higher-quality multi-view pairs.

\begin{figure*}[t]
    \centering
    \includegraphics[width=1.0\linewidth]{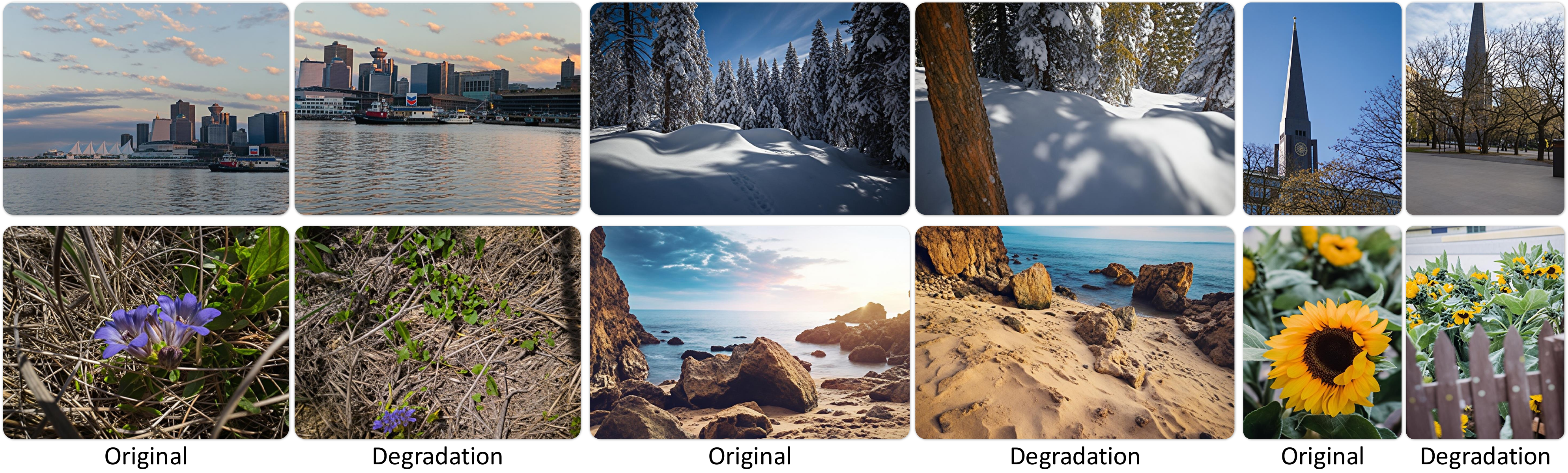}
    \vspace{-15pt}
    \caption{Degradation examples generated by our degradation model.}
    \label{supp:fig:degradation}
    \vspace{-5pt}
\end{figure*}

\begin{figure*}[t]
    \centering
    \includegraphics[width=1.0\linewidth]{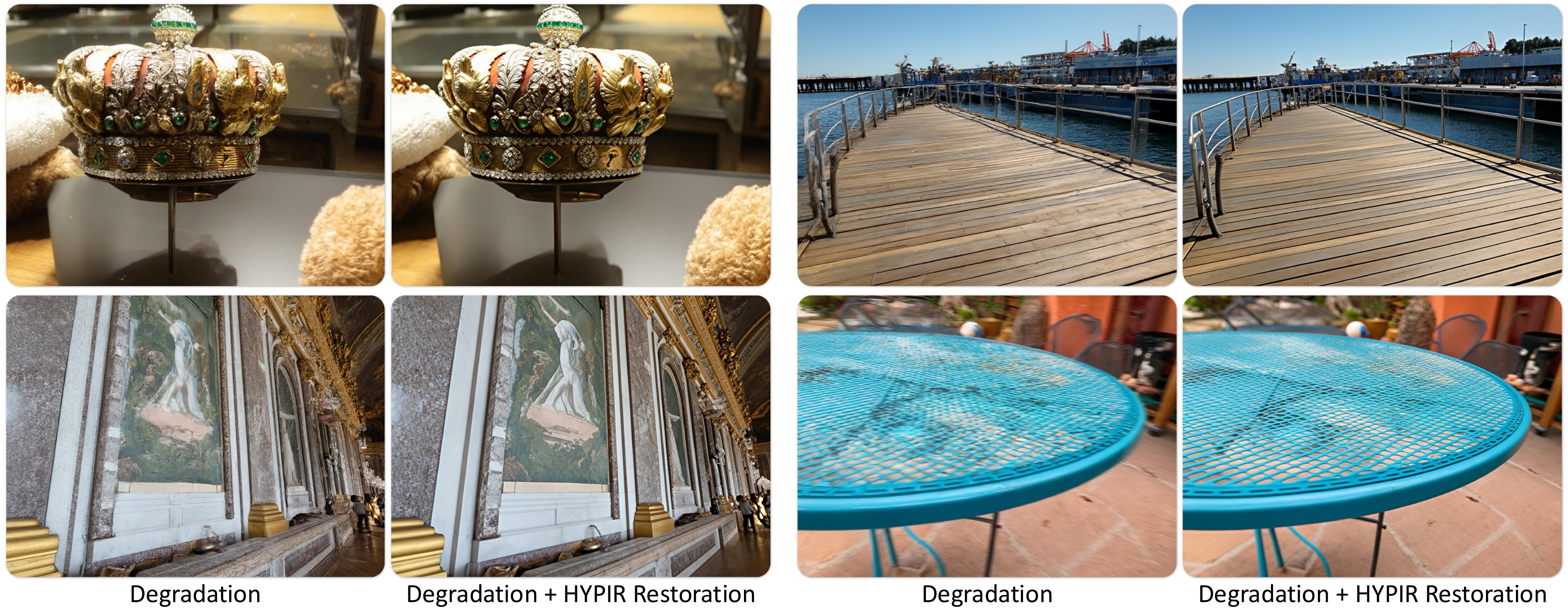}
    \vspace{-15pt}
    \caption{We use a state-of-the-art restoration model HYPIR to enhance the quality of degraded images.}
    \label{supp:fig:hypir_restore}
    \vspace{-10pt}
\end{figure*}

\vspace{2pt}\noindent\textbf{Text–vision joint training of the degradation model}. 
%
As illustrated in Fig.~\textcolor{red}{11} of the main paper, 
incorporating text guidance leads to higher-quality results compared with image-only training. 
Therefore, we adopt the same text–vision joint training strategy for the degradation model. 
Let the good image be denoted as \texttt{I\_good}, the corresponding poor image as \texttt{I\_poor}, the task prompt as \texttt{T\_task}, the predicted text guidance as \texttt{T\_guide}, and the degradation model as \texttt{f()}. 
The degradation model is formulated as: \texttt{I\_poor, T\_guide = f(I\_good, T\_task)}.

To train such a degradation model, we need to construct \texttt{<T\_task,I\_poor,I\_good,T\_guide>} samples. 
The image pairs \texttt{<I\_poor, I\_good>} are collected from multi-view datasets, as illustrated 
in Fig.~\textcolor{red}{5} of the main paper. 
The task prompt \texttt{T\_task}, analogous to the three sub-tasks, is randomly sampled from a set of predefined template sentences such as ``Change the viewpoint to \textit{worsen} the composition''. 
Finally, we input each image pair into the vision language model Qwen2.5-VL-32B~\cite{qwen2.5_vl} to annotate the text guidance \texttt{T\_guide}, which describes how the good image is transformed into the poor one.

With the above collected data, we could finally train the degradation model using the same text–vision joint training strategy described 
in Sec.~\textcolor{red}{4} of the main paper.
Some qualitative degradation examples generated by our degradation model are provided in \cref{supp:fig:degradation}.

\vspace{2pt}\noindent\textbf{Image restoration to enhance degraded outputs}. 
As stated in the main paper, we apply the degradation model to human-taken photos produces poor-composition images. 
However, these degraded outputs may occasionally suffer from low visual quality with noticeable artifacts, as shown in \cref{supp:fig:hypir_restore}. 
To mitigate this issue, we employ the state-of-the-art image restoration model HYPIR~\cite{hypir} to enhance the visual quality of degraded images. 
Specifically, we first down-sample each image by a factor of 4 and then use HYPIR to up-sample it by the same factor, which further suppresses generative artifacts. 
As illustrated in \cref{supp:fig:hypir_restore}, HYPIR substantially improves image fidelity, yielding cleaner and more natural-looking degraded images.

\subsection{Dataset Statistics}
\label{supp:subsec:statis}

\textbf{Image statistics}. We have provided a rough statistic of our collected dataset 
in the Tab.~\textcolor{red}{1} of the main paper. 
Here we perform a more detailed statistic of each sub-task in \cref{supp:tab:statis_shift,supp:tab:statis_zoomin,supp:tab:statis_view}. 
Since the shift and zoom-in pairs are sampled from the original images in existing cropping datasets including GAIC~\cite{gaic_v2}, CPC~\cite{cpc}, SACD~\cite{sacnet_sacd}, FLMS~\cite{flms}, FlickrCrop~\cite{flickr}, CUHKCrop~\cite{cuhkcrop}, both the original images and pairs are included in the statistics. 
In \cref{supp:tab:statis_shift}, besides the collected shift pairs from GAIC and CPC, we also collect a small number of shift pairs from FlickrCrop and other resources (\eg, AIGC creation), denoted as ``Other''.

\begin{table}[t]
    \centering
    \footnotesize
    \setlength\tabcolsep{11pt}
    \caption{Statistics of training / test shift dataset.}
    \vspace{-5pt}
    \begin{tabular}{c|ccc}
    \toprule
    & GAIC & CPC & Other \\
    \midrule
    \# Original & 2,613 / 277 & 6,128 / 682 & 621 / - \\
    \# Pairs & 72,710 / 4,578 & 80,486 / 6,432 & 698 / - \\
    \bottomrule
    \end{tabular}
    \label{supp:tab:statis_shift}
\vfill
\vspace{8pt}
    \centering
    \scriptsize
    \setlength\tabcolsep{2.8pt}
    \caption{Statistics of training / test zoom-in dataset.}
    \vspace{-5pt}
    \begin{tabular}{c|cccccc}
    \toprule
    & GAIC & CPC & SACD & FLMS & FlickrCrop & CUHKCrop \\
    \midrule
    \# Original & 1,603 / 147 & 4,099 / 459 & 796 / 70 & 338 / - & 110 / - & 43 / - \\
    \# Pairs & 3,830 / 327 & 7,426 / 869 & 813 / 70 & 686 / - & 110 / - & 51 / - \\
    \bottomrule
    \end{tabular}
    \label{supp:tab:statis_zoomin}
\vfill
\vspace{8pt}
    \centering
    \footnotesize
    \setlength\tabcolsep{10pt}
    \caption{Statistics of training / test view-change dataset.}
    \vspace{-5pt}
    \begin{tabular}{c|ccc}
    \toprule
    & DL3DV & Unsplash Lite & Our Collected \\
    \midrule
    \# Pairs & 13,335 / 710 & 7,289 / 1,000 & 4,340 / 719 \\
    \bottomrule
    \end{tabular}
    \label{supp:tab:statis_view}
    \vspace{-5pt}
\end{table}

\begin{figure}
    \centering
    \includegraphics[width=1.0\linewidth]{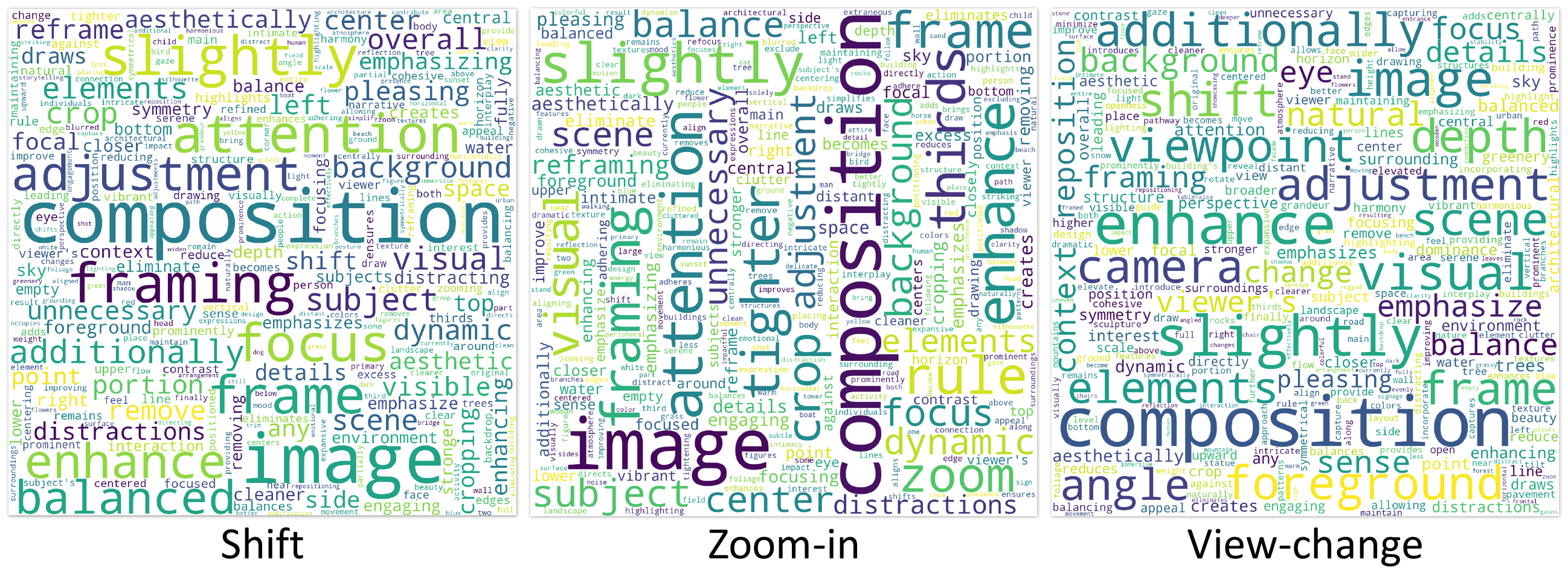}
    \vspace{-15pt}
    \caption{Word cloud of the text guidance.}
    \label{supp:fig:wordcloud}
    \vspace{-10pt}
\end{figure}

\vspace{2pt}\noindent\textbf{Text statistics}. We perform statistics on the text guidance length of three sub-tasks in \cref{supp:tab:statis_text}. 
The average word length is about 100 words, providing sufficiently rich yet concise guidance for composition refinement. 
The word cloud of the text guidance is depicted in \cref{supp:fig:wordcloud}, where high-frequency words such as ``composition'', ``balance'', ``framing'', ``angle'', ``viewpoint'', ``tighter'', and ``depth'' are all closely related to compositional principles.

\begin{table}[t]
    \centering
    \footnotesize
    \setlength\tabcolsep{11pt}
    \caption{Statistics of the text guidance length.}
    \vspace{-9pt}
    \begin{tabular}{c|ccc}
    \toprule
    & Shift & Zoom-in & View-change \\
    \midrule
    \# Word Length & 94.65 & 102.60 & 97.06 \\
    \# String Length & 613.43 & 663.16 & 632.77 \\
    \bottomrule
    \end{tabular}
    \label{supp:tab:statis_text}
    \vspace{-5pt}
\end{table}

\subsection{Other Details}
\label{supp:subsec:task_prompt}

\textbf{Task prompt}. The task prompts for all three sub-tasks are provided in \cref{supp:tab:prompt_each_task}. 
Moreover, the prompts used for the two auto tasks are listed in \cref{supp:tab:prompt_auto_task}.

\section{Limitations and Discussions}

\begin{figure}[t]
    \centering
    \includegraphics[width=1.0\linewidth]{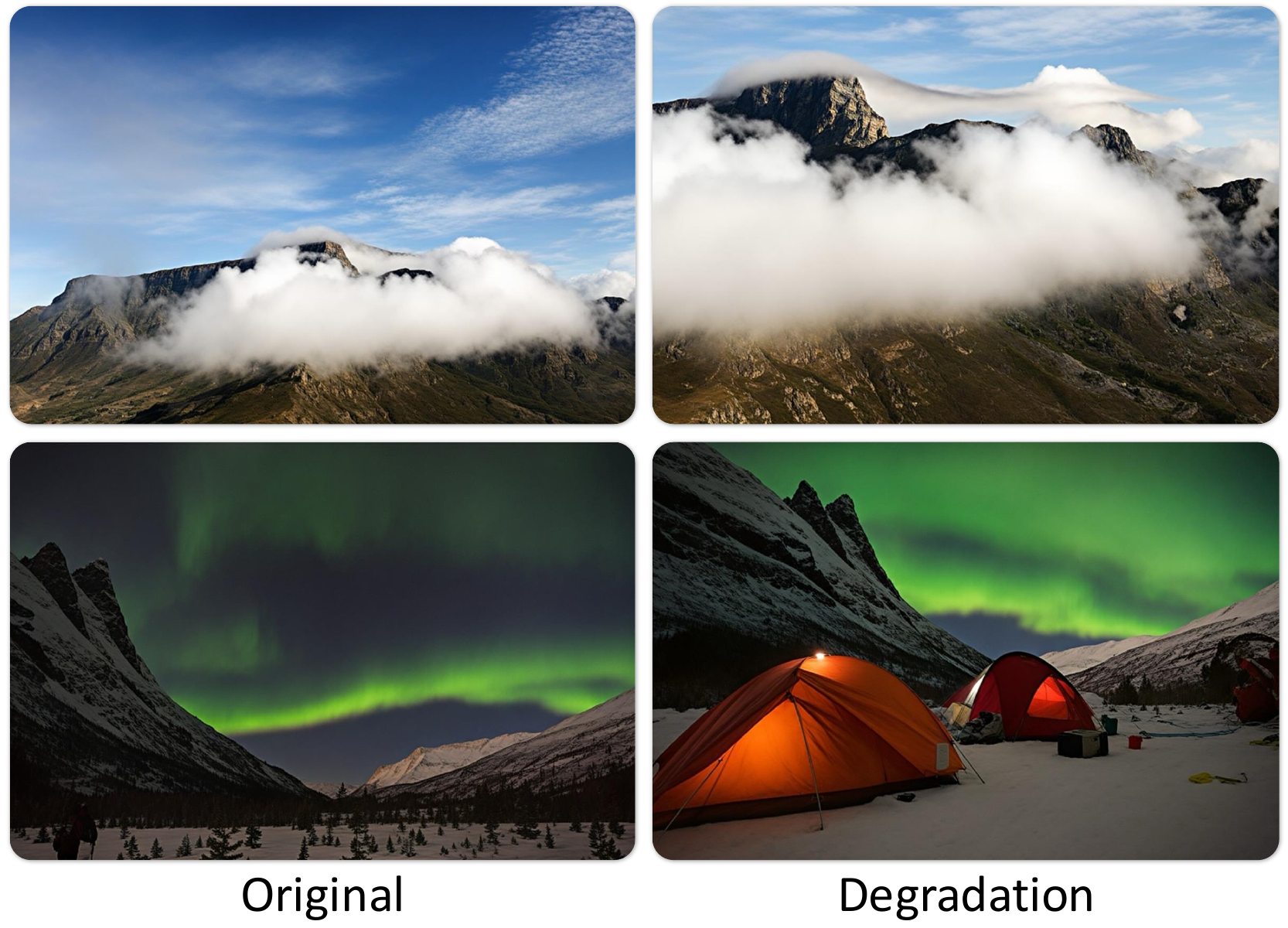}
    \vspace{-15pt}
    \caption{The degradation model may fail to produce a poorly composed image when given an expert-taken photo, in which case the degraded output can still appear well-composed.}
    \label{supp:fig:good_good}
    \vspace{-5pt}
\end{figure}

First, the degradation model may fail to worsen the good composition. 
This issue is particularly common for expert-taken photos in the Unsplash Lite~\cite{unsplash_lite} dataset, where the degraded outputs can remain well-composed, as illustrated in \cref{supp:fig:good_good}. 
Fortunately, we observe empirically that including such \texttt{<good,good>} pairs does not harm the model performance. 
Moreover, since the users may also input such already well-composed images, we intentionally retain these samples rather than filtering them out.

\begin{figure}[t]
    \centering
    \includegraphics[width=1.0\linewidth]{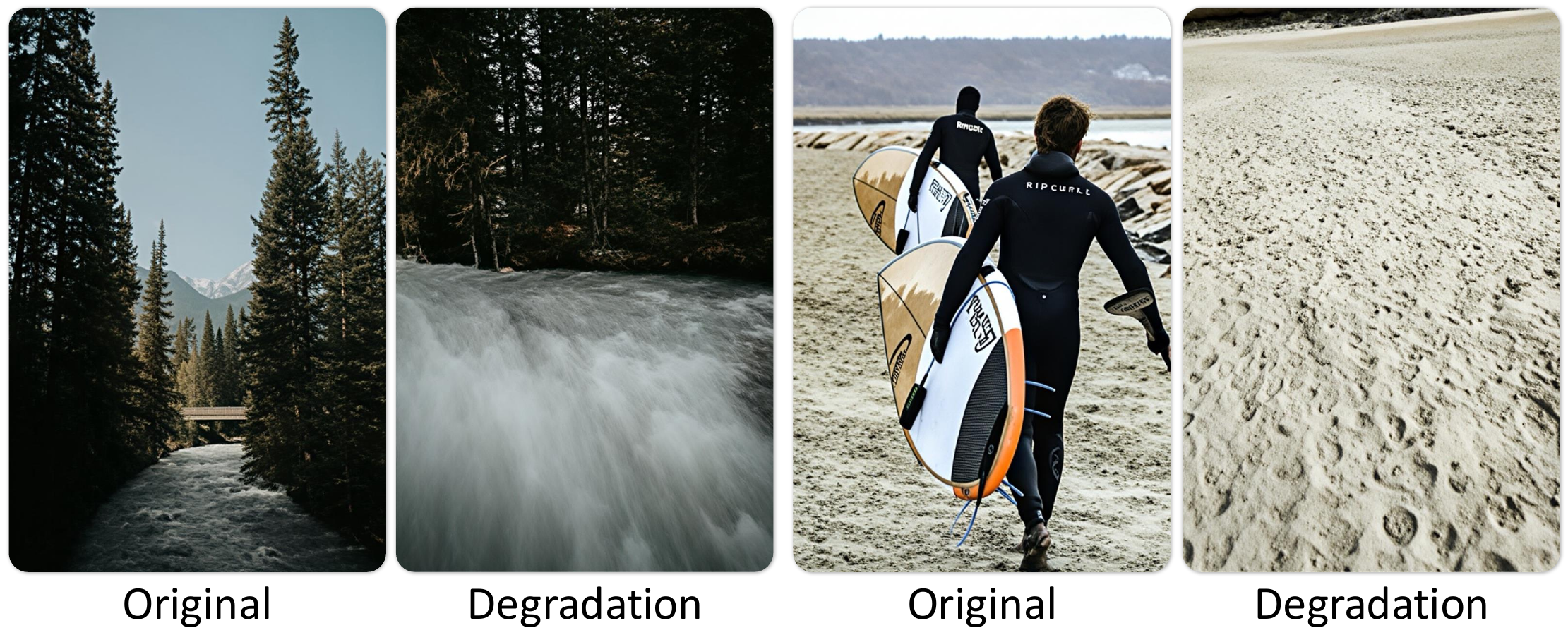}
    \vspace{-15pt}
    \caption{The degradation model may alter the image excessively, leading to semantically inconsistent pairs.}
    \label{supp:fig:inconsistency}
    \vspace{-10pt}
\end{figure}

Second, the degradation model may change the image excessively, producing pairs that are no longer semantically consistent. As shown in \cref{supp:fig:inconsistency}, the degraded output deviates notably from the original content. Although we perform data filtering, a small number of such inconsistent pairs may still remain. Future work should explore a more controllable degradation model to reduce inconsistency.

Third, the generated images may exhibit relatively low visual quality, as depicted in \cref{supp:fig:hypir_restore}. However, image quality is not the primary focus of our work, and we are concerned chiefly with composition. Improving the quality of generated images can be explored in our future work.

\begin{figure}
    \centering
    \includegraphics[width=1.0\linewidth]{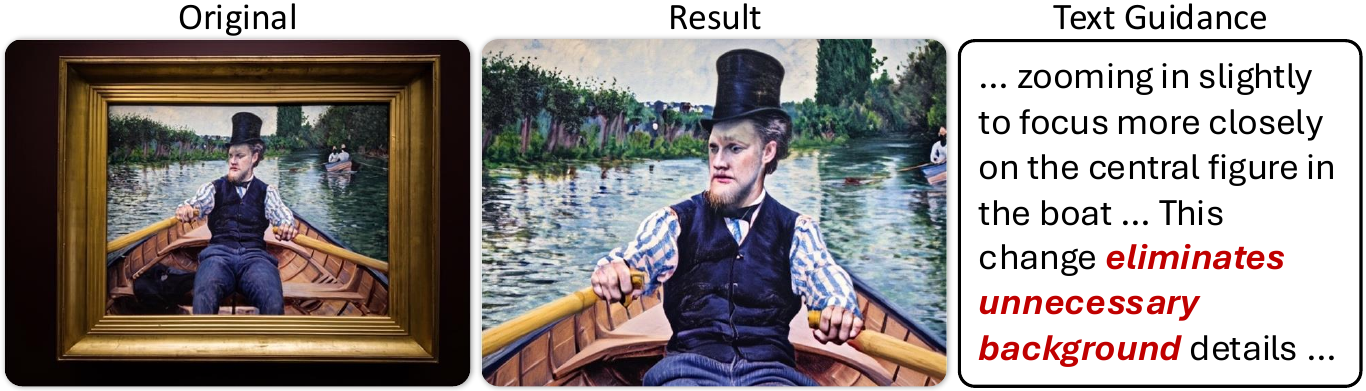}
    \vspace{-18pt}
    \caption{A failure case where our \method misinterprets a painting as a real photograph and attempts to refine the composition by removing the ``background'' (\ie, the painting frame).}
    \label{fig:fail}
    \vspace{-16pt}
\end{figure}

Fourth, we emphasize that synthetic data provides only an initial foundation for training composition instruction models. While it enables large-scale supervision and controlled variation, it cannot fully replace real-world images captured by everyday mobile users, which exhibit diverse shooting habits, device characteristics, and natural composition errors. Constructing such a real, human-captured dataset will be crucial for advancing composition instruction and remains an important direction for future work.
We present a failure case in \cref{fig:fail} where \method\ misinterprets a painting as a real photograph and attempts to refine the composition by removing the perceived ``background'' (\ie, the painting frame). 
This demonstrates the limitations of our small-scale datasets.

Finally, although it is best to deploy the model on a robot (like~\cite{autophoto}) to automatically capture photos, real-world robot deployment remains challenging due to the compute and latency constraints of running a 14B model on-device. Building a reliable hardware-in-the-loop platform is beyond the scope of this paper and is left for future work.

\begin{table*}[t]
    \centering
    \footnotesize
    \caption{\textbf{Task prompt} of three sub-tasks.}
    \vspace{-5pt}
    \begin{tabular}{ll}
    \toprule
    \# & \multicolumn{1}{c}{Shift Prompt} \\
    \midrule
    1 & Shift the framing of this scene to improve its overall composition. \\
    2 & Shift the camera to make the scene more visually appealing. \\
    3 & Adjust the framing by shifting to create a stronger composition. \\
    4 & Reframe by shifting to enhance the visual composition. \\
    5 & Shift the view to improve the image composition. \\
    6 & Reframe by shifting to make the scene more pleasing. \\
    7 & Refine the image composition by shifting the frame. \\
    8 & Shift the scene to enhance the composition. \\
    9 & Enhance the scene by shifting the camera framing. \\
    10 & Shift the frame to create a stronger composition. \\
    \midrule
    \# & \multicolumn{1}{c}{Zoom-in Prompt} \\
    \midrule
    1 & Zoom in the framing of this scene to improve its overall composition. \\
    2 & Tighten the framing by zooming in to make the image more appealing. \\
    3 & Zoom in to refine the composition and enhance the visual effect. \\
    4 & Narrow the framing by zooming in for a more pleasing result. \\
    5 & Adjust the composition by zooming in to strengthen the image. \\
    6 & Reframe by zooming in to make the scene look more polished. \\
    7 & Refine the composition by zooming in to simplify the frame. \\
    8 & Enhance the image by narrowing the view with a zoom-in. \\
    9 & Strengthen the composition by zooming in. \\
    10 & Make the image look cleaner by zooming in on the framing. \\
    \midrule
    \# & \multicolumn{1}{c}{View Change Prompt} \\
    \midrule
    1 & Change the viewpoint of this scene to improve its overall composition. \\
    2 & Make a view change to find a better shooting point and improve the composition. \\
    3 & Perform a view change to choose a more appealing shooting spot. \\
    4 & Select a new shooting view to create a better composition. \\
    5 & Change the shooting view to achieve a more visually pleasing shooting point. \\
    6 & Explore a different shooting spot to improve the overall composition. \\
    7 & Change the shooting view to create a more effective composition. \\
    8 & Make a view change to enhance the composition of the scene framing. \\
    9 & Apply a view change for a more engaging and visually strong composition. \\
    10 & Change the viewpoint to make the composition more attractive. \\
    \bottomrule
    \end{tabular}
    \label{supp:tab:prompt_each_task}
\vfill
\vspace{20pt}
    \centering
    \footnotesize
    \caption{\textbf{Task prompt} of two auto tasks.}
    \vspace{-5pt}
    \begin{tabular}{ll}
    \toprule
    \# & \multicolumn{1}{c}{Prompt for Static Auto Task (Shift or Zoom-in)} \\
    \midrule
    1 & Refine the composition through shift or zoom-in adjustments. \\
    2 & Improve the composition by combining shift or zoom-in operations. \\
    3 & Enhance the image composition with coordinated shift or zoom-in refinement. \\
    4 & Adjust the framing through shift or zoom-in to achieve a better composition. \\
    5 & Make the composition more appealing using shift or zoom-in adjustments. \\
    6 & Refine the scene composition with gentle shift or zoom-in movement. \\
    7 & Improve the framing by applying continuous shift or zoom-in optimization. \\
    8 & Enhance the overall composition through integrated shift or zoom-in refinement. \\
    9 & Adjust the frame smoothly using shift or zoom-in to strengthen composition. \\
    10 & Refine the image framing through natural shift or zoom-in enhancement. \\
    \midrule
    \# & \multicolumn{1}{c}{Prompt for Full Auto Task (All Three Tasks)} \\
    \midrule
    1 & Capture this scene with better composition. \\
    2 & Enhance the composition of this scene. \\
    3 & Refine the composition to make this scene more visually pleasing. \\
    4 & Adjust the image to improve its composition. \\
    5 & Reframe this scene to achieve a more pleasing visual composition. \\
    6 & Refine the framing through shift, zoom-in, or view change. \\
    7 & Improve the composition through shift, zoom-in, or viewpoint adjustment. \\
    8 & Enhance the image composition through shift, zoom-in, and view change. \\
    9 & Optimize the composition through shift, zoom-in, or new viewpoint exploration. \\
    10 & Refine the scene composition through shift, zoom-in, or view change as needed. \\
    \bottomrule
    \end{tabular}
    \label{supp:tab:prompt_auto_task}
\end{table*}

\definecolor{myblue}{rgb}{0.18, 0.43, 0.73}
\definecolor{myred}{rgb}{0.75, 0., 0.}
\begin{figure*}[t]
    \centering
    \includegraphics[width=0.9\linewidth]{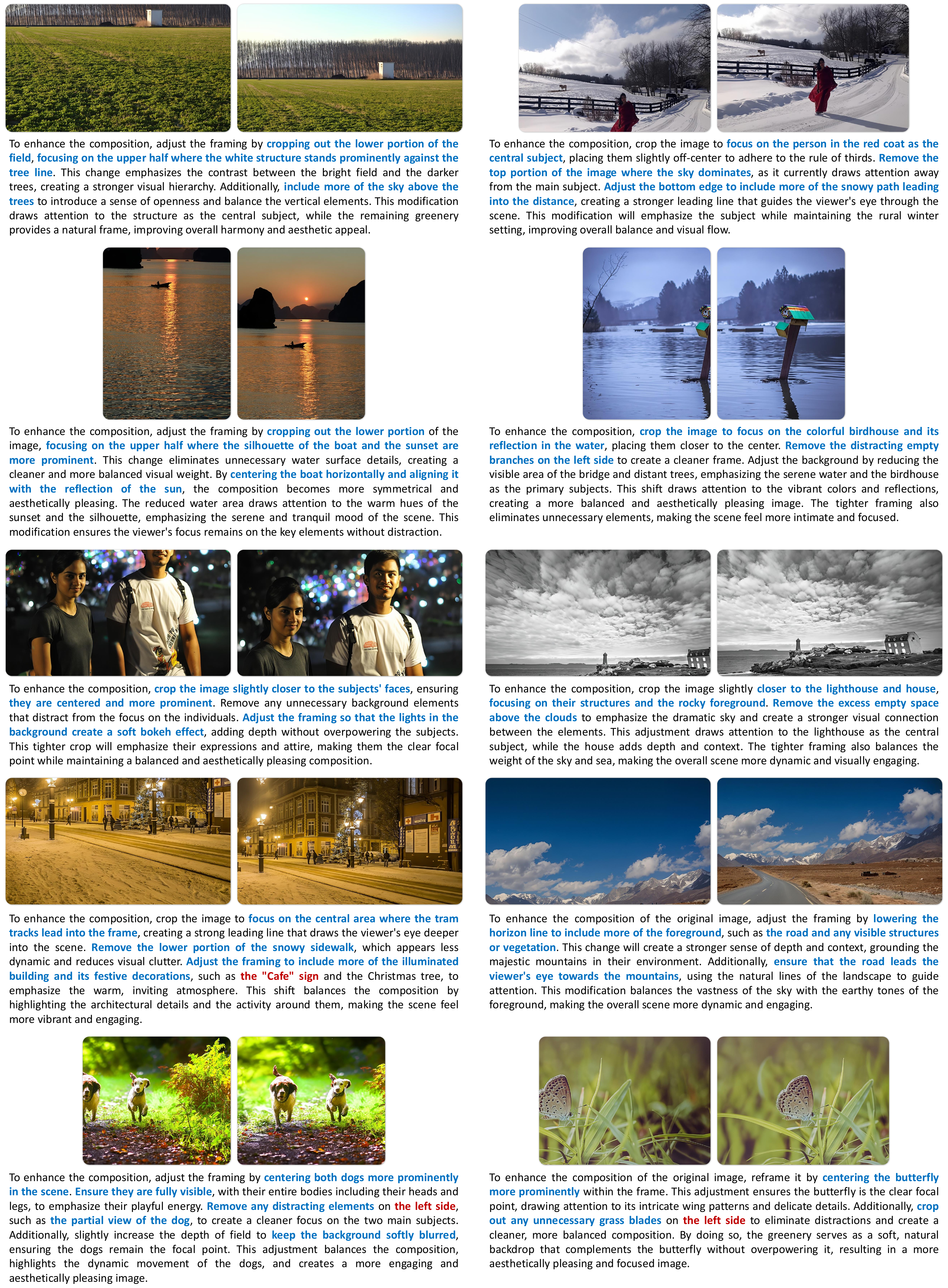}
    \vspace{-5pt}
    \caption{Qualitative results of the shift task. Left / right / bottom: original image / generated example image / text guidance. Actionable textual suggestions are highlighted in \textbf{\textcolor{myblue}{blue}}, while hallucinated descriptions are marked in \textbf{\textcolor{myred}{red}}.
    }
    \label{supp:fig:res_shift}
\end{figure*}

\definecolor{myblue}{rgb}{0.18, 0.43, 0.73}
\definecolor{myred}{rgb}{0.75, 0., 0.}
\begin{figure*}[t]
    \centering
    \includegraphics[width=0.9\linewidth]{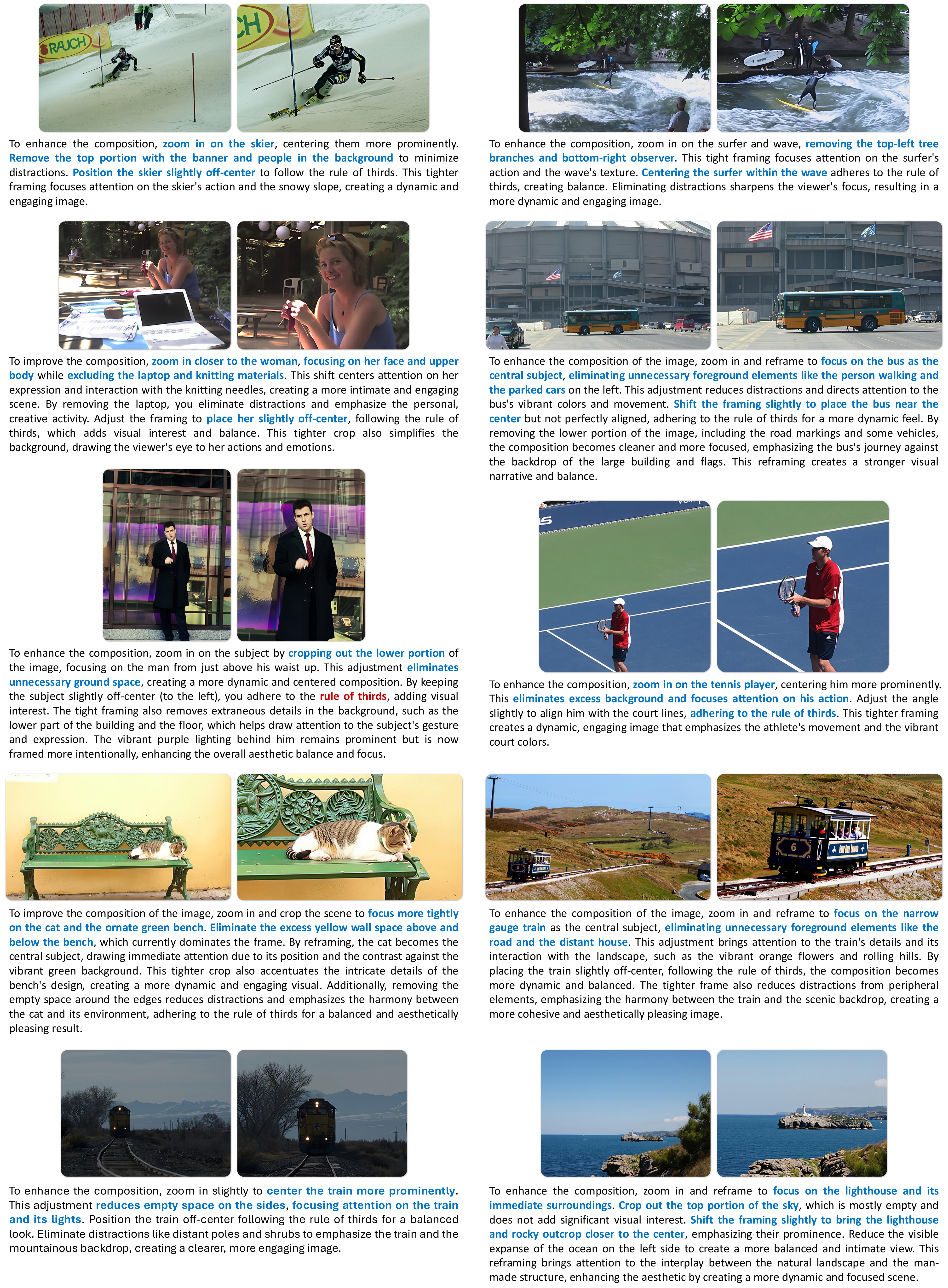}
    \vspace{-5pt}
    \caption{Qualitative results of the zoom-in task. Left / right / bottom: original image / generated example image / text guidance. Actionable textual suggestions are highlighted in \textbf{\textcolor{myblue}{blue}}, while hallucinated descriptions are marked in \textbf{\textcolor{myred}{red}}.
    }
    \label{supp:fig:res_zoomin}
\end{figure*}

\definecolor{myblue}{rgb}{0.18, 0.43, 0.73}
\definecolor{myred}{rgb}{0.75, 0., 0.}
\begin{figure*}[t]
    \centering
    \includegraphics[width=0.9\linewidth]{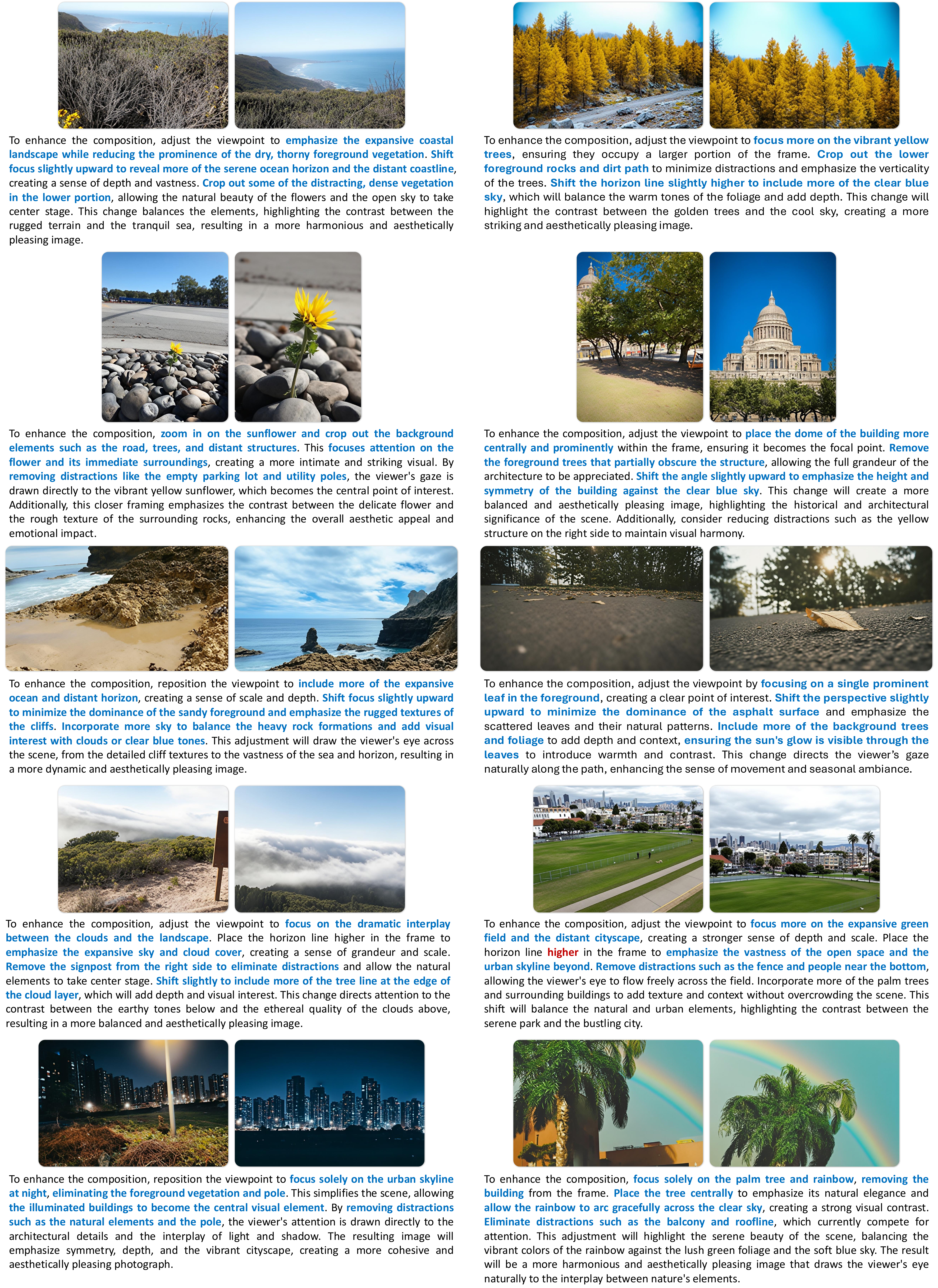}
    \vspace{-5pt}
    \caption{Qualitative results of the view-change task. Left / right / bottom: original image / generated example image / text guidance. Actionable textual suggestions are highlighted in \textbf{\textcolor{myblue}{blue}}, while hallucinated descriptions are marked in \textbf{\textcolor{myred}{red}}.
    }
    \label{supp:fig:res_view}
\end{figure*}

\end{document}